%% file: main.tex
\documentclass[journal]{IEEEtran}
\usepackage{amsmath,amsfonts}
\usepackage{algorithmic}
\usepackage{algorithm}
\usepackage{array}
\usepackage[caption=false,font=normalsize,labelfont=sf,textfont=sf]{subfig}
\usepackage{textcomp}
\usepackage{stfloats}
\usepackage{url}
\usepackage{verbatim}
\usepackage{graphicx}
\usepackage{cite}
\usepackage{xcolor}
\usepackage{booktabs}
\usepackage{multirow}
\usepackage{mathrsfs}
\usepackage{amssymb}
\newcommand{\emphasize}[1]{\textbf{\underline{#1}}}

\begin{document}
\bstctlcite{IEEEtranBSTCTL}

\title{Adaptive Linear Networks for Irregular Multivariate Time Series Forecasting in Big Data Systems}

\author{

    Chengsen~Wang,
    Qi~Qi,~\IEEEmembership{Senior~Member,~IEEE},
    Bo~He,~\IEEEmembership{Member,~IEEE},\\
    Haifeng~Sun,~\IEEEmembership{Member,~IEEE},
    Zirui~Zhuang,~\IEEEmembership{Member,~IEEE},
    Jingyu~Wang,~\IEEEmembership{Senior~Member,~IEEE}

    \thanks{
        Chengsen~Wang, Qi~Qi, Bo~He, Haifeng~Sun, Zirui~Zhuang and Jingyu~Wang are with the State Key Laboratory of Networking and Switching Technology, Beijing University of Posts and Telecommunications, Beijing, China. (email: \{cswang, qiqi8266, hebo, hfsun, zhuangzirui, wangjingyu\}@bupt.edu.cn)
    }
    \thanks{
        Chengsen~Wang and Qi~Qi contribute equally to this work. Jingyu~Wang is the corresponding author.
    }

}

\markboth{Journal of \LaTeX\ Class Files,~Vol.~14, No.~8, August~2021}%
{Shell \MakeLowercase{\textit{et al.}}: A Sample Article Using IEEEtran.cls for IEEE Journals}

\maketitle

\begin{abstract}

   Large-scale sensing, clinical, transportation, energy, and climate systems often produce irregular multivariate time series with sparse observations, heterogeneous sampling intervals, and asynchronous variables. Forecasting at externally supplied future query coordinates requires a model to accommodate variable observation lengths and time locations while retaining efficient variable interaction. We present AiT (\emphasize{A}daptive \emphasize{i}nverted \emphasize{T}ransformer), whose ALinear (\emphasize{A}daptive \emphasize{Linear}) module generates sample-adaptive temporal weights from observation or query coordinates. More precisely, ALinear is time-conditioned attention pooling without value projection. Available observation or query coordinates generate the corresponding keys or queries, the fixed-shape counterpart may use learned default slots, and the unprojected inputs serve as values. The name ``Linear'' therefore denotes only the final weighted-sum output form and not equivalence to an unrestricted affine layer or to DLinear. AiT uses ALinear to encode variable-length histories and decode query-conditioned forecasts, and applies Transformer blocks over the resulting variable tokens to model cross-variable relationships. The evaluation on four public irregular forecasting datasets examines prediction accuracy and computational efficiency, while controlled protocols are specified for temporal-design, scalability, missingness, and statistical analyses.

\end{abstract}



\section{Introduction}
\label{sec:Introduction}

Time series forecasting is a core task in big data analytics for finance, transportation, energy, healthcare, and climate services \cite{TBD_DynamicHypergraph, TBD_LightST, TBD_WeatherTF}. Modern forecasting models include statistical methods \cite{ARIMA}, RNNs (Recurrent Neural Networks) \cite{DeepAR, LSTM}, CNNs (Convolutional Neural Networks) \cite{TimesNet, MICN}, GNNs (Graph Neural Networks) \cite{MTGNN, FourierGNN}, and Transformers \cite{Informer, Crossformer}. On regularly sampled data, static linear forecasters such as DLinear \cite{DLinear} offer a strong accuracy-efficiency trade-off, while iTransformer \cite{iTransformer} combines a fixed-length temporal projection with attention over variable tokens.

Most multivariate forecasting models assume RMTS (Regular Multivariate Time Series), where observations lie on a fixed grid and variables share an index-to-time mapping. Clinical records, urban sensors, and climate stations instead commonly produce IMTS (Irregular Multivariate Time Series), in which each variable can have its own observation times and missing pattern \cite{IMTS}. Existing IMTS approaches include interpolation and time-aware attention \cite{IPN, mTAND}, continuous-time dynamics \cite{Latent_ODE, Neural_Flows, CRU, TBD_NeuralRDE}, and recent probabilistic, graph, patch, and Transformer forecasters \cite{ProFITi, GraFITi, T_PatchGNN, ISTS_PLM, Hi_Patch}. These developments motivate a narrower question. Can variable-length histories and externally specified future queries be connected by a constrained temporal aggregator while cross-variable modeling is deferred to a fixed-size latent space?

\begin{figure}
    \centering
    \resizebox{0.95\linewidth}{!}
    {
        \includegraphics{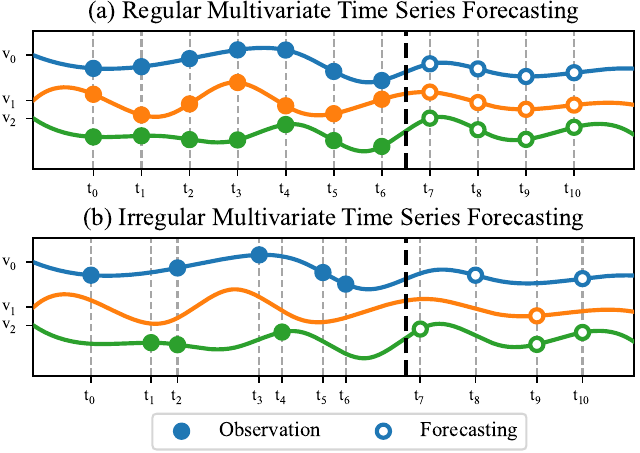}
    }
    \caption{The intuitive comparison of regular multivariate time series forecasting and irregular multivariate time series forecasting.}
    \label{fig:01}
\end{figure}

As illustrated in Figure \ref{fig:01}, IMTS forecasting differs from regular-grid forecasting in two relevant respects. First, variable observation counts and coordinates differ across samples, so a single static temporal matrix can incur shape mismatch or attach the same index weight to different physical times. Second, asynchronous variables are difficult to compare at the raw-event level because naive alignment mixes observed values with placeholders. AiT addresses these issues by separating per-variable, coordinate-conditioned temporal aggregation from interaction among fixed-length variable representations.

We propose AiT (\emphasize{A}daptive \emphasize{i}nverted \emphasize{T}ransformer), comprising a Temporal Encoder, a Spatial Encoder, and a Predictor. Its core ALinear module computes coordinate-dependent query-key scores and applies the resulting normalized weights directly to unprojected input values. Thus, ALinear is best characterized as time-conditioned attention pooling without value projection, rather than as a standard Linear layer or a claim that attention itself is new. The design contribution is the constrained use of coordinate-sourced weights as an adaptive replacement for a fixed temporal projection in both variable-length history encoding and query-conditioned decoding. After each series has incorporated its own timestamps and mask, the Spatial Encoder applies Transformer blocks to variable tokens.

The contributions of this paper are summarized as follows:

\begin{itemize}

    \item We formulate ALinear as time-conditioned attention pooling without value projection. Available temporal coordinates generate the corresponding query or key side, learned defaults represent a fixed-shape counterpart when needed, and the unprojected inputs provide the values.

    \item We develop AiT, which uses this constrained aggregator to unify variable-length history encoding and externally query-conditioned prediction, followed by latent-space variable interaction.

    \item We evaluate the accuracy-efficiency trade-off through a composite literature benchmark and specify controlled protocols for temporal aggregation, paired significance, spatial scalability, and escalated-MCAR robustness.

\end{itemize}

\section{Related Work}
\label{sec:Related_Work}

\subsection{Regular Multivariate Time Series Analysis}
\label{subsec:Regular_Multivariate_Time_Series_Analysis}

RMTS forecasting assumes that every variable is observed on a common time grid. Under this assumption, statistical models such as ARIMA \cite{ARIMA} and neural sequence models such as RNNs \cite{DeepAR, LSTM} process time steps in a fixed order. CNN-based models \cite{TimesNet, MICN} and attention-based models \cite{PatchTST, Crossformer} improve long-range temporal modeling. GNN-based methods model variable relationships through graph structures \cite{StemGNN, CrossGNN, FourierGNN}. Recent TBD studies also indicate that dynamic graph, hypergraph, and spatio-temporal attention structures are useful for multivariate forecasting and traffic or weather prediction in data-intensive systems \cite{TBD_DynamicHypergraph, TBD_DecoupledCausal, TBD_DTWformer, TBD_LightST, TBD_WeatherTF}.

Linear forecasting networks such as DLinear \cite{DLinear} use a learned, static temporal matrix whose rows are shared across samples and whose columns assume a fixed input length and index-to-time mapping. This static projection is efficient on regular grids, but it does not natively adapt when observation counts or coordinates change. At the other end, vanilla temporal self- or cross-attention \cite{Transformer} normally constructs temporal tokens from observed content and positional or timestamp features, independently projects queries, keys, and values, and may permit pairwise interactions among historical time points. iTransformer \cite{iTransformer} instead embeds each complete variable series as a token and uses attention primarily for cross-variable interaction. Recent work further studies when channel interaction is beneficial \cite{CI, LIFT} and how higher-order, lead-lag, or multi-scale structures can refine variable dependencies \cite{STHyper, MillGNN}. ALinear occupies a deliberately constrained point between static temporal projection and full temporal attention. Without a value projection, it generates normalized weights from temporal coordinates and applies them directly to the raw values. These designs differ mainly in weight normalization, adaptivity, and value handling. A static linear projection uses fixed, unnormalized weights and adapts to neither the sample nor the input length. Its score cost is $O(L_{in}L_{out})$. Vanilla temporal attention uses Softmax-normalized, sample-adaptive weights and projects the values, so it is value-conditioned. Its score cost is $O(L_{in}L_{out}D)$. ALinear keeps the normalization, sample adaptivity, and variable-length support of attention but applies its weights to raw unprojected values, so it is not value-conditioned. Its score cost is the same $O(L_{in}L_{out}D)$.

\subsection{Irregular Multivariate Time Series Analysis}
\label{subsec:Irregular_Multivariate_Time_Series_Analysis}

IMTS analysis spans classification, imputation, and forecasting. GRU-D \cite{GRU_D} incorporates missingness through decay and imputation, whereas set and graph encoders \cite{SeFT, RainDrop, Warpformer} represent irregular observations without requiring a fully observed grid. Interpolation-Prediction Networks \cite{IPN} learn interpolation functions, and mTAND \cite{mTAND} performs multi-time attention from fixed reference queries to temporal tokens with projected value features. These time-aware designs establish that temporal coordinates can guide aggregation, but their objectives and Q/K/V constructions differ from the direct raw-value pooling and query-conditioned forecasting of ALinear.

For IMTS forecasting, continuous-time models such as Latent-ODE \cite{Latent_ODE}, Neural Flows \cite{Neural_Flows}, CRU \cite{CRU}, and neural differential equation variants \cite{TBD_NeuralRDE} model dynamics between observations. ProFITi \cite{ProFITi} uses conditional normalizing flows for probabilistic forecasts, GraFITi \cite{GraFITi} constructs graphs over irregular samples, T-PatchGNN \cite{T_PatchGNN} combines transformable temporal patches with graph modeling, ISTS-PLM \cite{ISTS_PLM} uses Transformer-based temporal and variable representations, and Hi-Patch \cite{Hi_Patch} builds hierarchical patch graphs. These methods commonly evaluate values at supplied future query coordinates, but they differ in their temporal representation, probabilistic objective, or interaction structure. AiT focuses on point forecasting and separates coordinate-conditioned, per-variable aggregation from Transformer interaction over fixed-length variable tokens.

\section{Preliminary}
\label{sec:Preliminary}

\subsection{Problem Definition}
\label{subsec:Problem_Definition}

\textbf{Definition 1 (Irregular Multivariate Time Series).} An IMTS with $N$ variables can be represented as $\mathcal{O}=\left\{\mathbf{o}_{1:L_n}^n\right\}_{n=1}^N=\left\{\left[\left(t_i^n, x_i^n\right)\right]_{i=1}^{L_n}\right\}_{n=1}^N$, where the $n$-th variable contains $L_n$ observations. The $i$-th observation of the $n$-th variable consists of the recorded time point $t_i^n$ and the corresponding value $x_i^n$. The time points $t_i^n$ are arranged chronologically, and the intervals between neighboring time points can vary. Due to irregular sampling intervals and missing values, different samples or variables may contain different numbers of observations within the same time span. Furthermore, the same observation index $i$ may correspond to different actual time points across samples or variables.

\textbf{Problem 1 (Query-Conditioned Irregular Multivariate Time Series Forecasting).} Given a historical IMTS $\mathcal{H}=\left\{\left[\left(t_i^n, x_i^n\right)\right]_{i=1}^{L_n}\right\}_{n=1}^N$ available at the forecast origin and externally supplied future forecasting coordinates $\mathcal{Q}=\left\{\left[q_j^n\right]_{j=1}^{Q_n}\right\}_{n=1}^N$, the task is to produce point estimates $\hat{\mathcal{X}}_{\mathcal{Q}}$ of the target values $\mathcal{X}_{\mathcal{Q}}$ conditioned on $\mathcal{H}$ and $\mathcal{Q}$. Under squared-error training, the population-optimal point estimate is the conditional mean $\mathbb{E}[\mathcal{X}_{\mathcal{Q}}\mid\mathcal{H},\mathcal{Q}]$. Each $q_j^n$ is a forecasting or evaluation request supplied by the application, not a future measurement event inferred from the realized observation process of the patient or sensor. The model receives neither the future values $\mathcal{X}_{\mathcal{Q}}$ nor the target-validity mask used to evaluate those values. This task is expressed as
\begin{equation}
    \mathcal{F}\!\left(\mathcal{H},\mathcal{Q}\right)\longrightarrow\hat{\mathcal{X}}_{\mathcal{Q}}.
    \label{eq:01}
\end{equation}
AiT therefore performs query-conditioned value forecasting rather than predicting the time, type, or observation intensity of the next event. Jointly forecasting event times and associated type or value marks is a temporal point-process problem, or a marked temporal point process (MTPP) when marks are included, and lies outside the task defined here.

\subsection{Canonical Pre-Alignment}
\label{subsec:Canonical_Pre_Alignment}

Canonical pre-alignment \cite{GRU_D} is an engineering representation used to batch IMTS samples with different observation times \cite{Latent_ODE, mTAND, Neural_Flows, T_PatchGNN}. It does not remove irregularity from the data or act as the forecasting model. Instead, it stores observed values, timestamps, and masks in fixed-shape tensors so that later modules can identify real observations. Specifically, IMTS data $\mathcal{O}$ is transformed into a triplet $\left(\mathcal{T}, \mathcal{X}, \mathcal{M}\right)$. First, $\mathcal{T} = \left[t_l\right]_{l=1}^{L} \in \mathbb{R}^{L}$ is obtained by merging all timestamps and sorting them, i.e., $\mathcal{T} = \bigcup_{n=1}^{N}\left\{t_i^{n}\right\}_{i=1}^{L_n}$. In practice, timestamps are scaled to the range $\left[0, 1\right]$ according to the total duration of both the look-back and forecast horizons. Second, the value matrix $\mathcal{X} = \left[\left[x_l^{n}\right]_{n=1}^{N}\right]_{l=1}^{L}\in\mathbb{R}^{L\times N}$ assigns $x_l^{n}=x_i^{n}$ if a matching observation exists, and otherwise stores a placeholder such as zero. Finally, the binary mask $\mathcal{M} = \left[\left[m_l^{n}\right]_{n=1}^{N}\right]_{l=1}^{L} \in \left\{0,1\right\}^{L\times N}$ has $m_l^{n}=1$ when $x_l^{n}$ is observed and $0$ otherwise. The timestamps and masks remain model inputs, so subsequent modules can use the original irregular sampling information rather than treating placeholders as observations.

\begin{figure*}
    \centering
    \resizebox{0.9\linewidth}{!}
    {
        \includegraphics{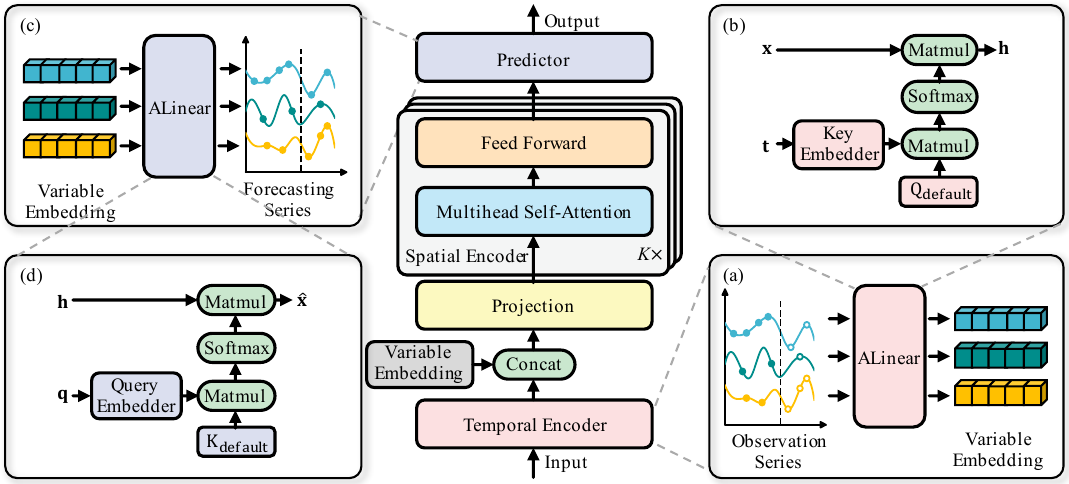}
    }
    \caption{The overall framework of AiT. Raw observation series from different variables are independently embedded into dynamic variable embeddings via the Temporal Encoder. The Projection then integrates dynamic and static variable embeddings. The aggregated variable embedding is subsequently fed into the Spatial Encoder to capture variable correlations. Finally, the Predictor individually transforms the embedding of each variable into the forecasting series.}
    \label{fig:02}
\end{figure*}

\section{Methodology}
\label{sec:Methodology}

\subsection{Framework Overview}
\label{subsec:Framework_Overview}

The overall architecture of AiT is illustrated in Figure \ref{fig:02} and comprises three components: a Temporal Encoder, a Spatial Encoder, and a Predictor. First, the Temporal Encoder models the irregularly sampled observation series $\mathbf{x}^n \in \mathbb{R}^{L_n}$ for each variable and embeds it into a fixed-length dynamic variable representation $\mathbf{h}^n_\text{dyna} \in \mathbb{R}^{D}$, where $D$ denotes the hidden dimension. Some variables may contain few or no observations in a sample. We therefore introduce a learnable static variable representation $\mathbf{h}^n_\text{stat} \in \mathbb{R}^{D}$ and concatenate it with $\mathbf{h}^n_\text{dyna}$ to provide variable-level prior information. This concatenated representation is projected back to $\mathbf{h}^n \in \mathbb{R}^{D}$ through an MLP. Next, the Spatial Encoder uses stacked Transformer blocks to model relationships among variable embeddings. This design captures variable interaction in latent space after each variable has encoded its own timestamps and masks. Finally, the Predictor receives the latent embedding of each variable and independently maps it to the forecasting series $\hat{\mathbf{x}}^n \in \mathbb{R}^{Q_n}$. We implement the Temporal Encoder and Predictor with ALinear to handle inconsistent historical observation times and future query times.

\subsection{Adaptive Linear}
\label{subsec:Adaptive_Linear}

In RMTS, a static temporal forecaster applies $\mathbf{y}=\mathbf{W}_{\mathrm{static}}\mathbf{x}$ with one index-bound matrix shared across samples. For irregular coordinates, ALinear instead applies $\mathbf{y}=\operatorname{Softmax}(\operatorname{Mask}(\mathbf{Q}(\mathbf{q})\mathbf{K}(\mathbf{t})^\mathsf{T},\mathbf{m}))\mathbf{x}$, so the weights vary with observation coordinates, output coordinates, and the validity mask. ALinear is therefore implemented by attention-style scoring and normalization and is most precisely described as time-conditioned attention pooling without value projection, and its name refers only to the terminal weighted sum.

Given the input data $\mathbf{x} \in \mathbb{R}^{L_{in}}$ and its associated time point $\mathbf{t} \in \mathbb{R}^{L_{in}}$, the objective of ALinear is to produce the value $\mathbf{y} \in \mathbb{R}^{L_{out}}$ corresponding to the output time point $\mathbf{q} \in \mathbb{R}^{L_{out}}$. In IMTS, the input length $L_{in}$ and the output length $L_{out}$ typically vary across samples or variables. ALinear first extracts high-dimensional features from each input time point in $\mathbf{t}$ and each output time point in $\mathbf{q}$ using the Key Embedder and Query Embedder. It converts them into keys $\mathbf{K} \in \mathbb{R}^{L_{in} \times D}$ and queries $\mathbf{Q} \in \mathbb{R}^{L_{out} \times D}$: \begin{equation}
    \begin{gathered}
        \begin{aligned}
            \mathbf{K} &= \operatorname{KeyEmbedder} \left( \mathbf{t} \right) \\
            \mathbf{Q} &= \operatorname{QueryEmbedder} \left( \mathbf{q} \right) \\
        \end{aligned}
    \end{gathered}
    \label{eq:02}
\end{equation}

To maintain architectural simplicity, both the Key Embedder and Query Embedder are constructed as an MLP with two linear transformations and a ReLU activation \cite{ReLU}. ALinear then derives the dynamic weight matrix $\mathbf{W} \in \mathbb{R}^{L_{out} \times L_{in}}$ by computing dot product scores between $\mathbf{Q}$ and $\mathbf{K}$. We apply Softmax \cite{Softmax} to normalize the weights along the input dimension. This normalization makes each output a convex weighted sum of observed input values. It also means that ALinear is not identical to a standard Linear layer, whose weights can be signed and unnormalized. To support batch processing for IMTS, we introduce placeholder values and a mask $\mathbf{m} \in \mathbb{R}^{L_{in}}$ during canonical pre-alignment. Placeholder values are excluded before Softmax normalization, following the masking operation used in attention mechanisms \cite{Transformer}. In the implementation, the mask removes invalid positions from the Softmax support, so the placeholder value is not treated as an observed zero. The entire process can be expressed as: \begin{equation}
    \begin{gathered}
        \begin{aligned}
            \mathbf{W} &= \operatorname{Softmax} \left( \operatorname{Mask} \left( \mathbf{Q}\mathbf{K}^\text{T}, \mathbf{m} \right) \right) \\
            \mathbf{y} &= \mathbf{W}\mathbf{x}
        \end{aligned}
    \end{gathered}
    \label{eq:03}
\end{equation}
If all entries in $\mathbf{m}$ are zero, the dynamic branch has no observed input values. In this case, the implementation bypasses the normalization over an empty support and returns an uninformative zero vector. The static variable representation in Equation \ref{eq:05} then provides the variable-level prior.

ALinear is related to time-aware attention and interpolation attention in prior IMTS models \cite{IPN, mTAND}. The main design difference is that ALinear does not introduce a value projection. It uses the raw input values as values and treats the time-conditioned attention weights as an adaptive replacement for a fixed temporal weight matrix. This design keeps the module simple, but also makes it different from a full attention block. Another practical difference is that ALinear uses learnable default query or key matrices when time points are absent or fixed, whereas mTAND relies on fixed reference time points for its multi-time attention. Because ALinear applies Softmax normalization, its weights are non-negative and row-normalized, so it should not be viewed as mathematically equivalent to a standard Linear layer with unrestricted signed weights. A more detailed comparison between ALinear and standard Linear, together with the algorithmic flow, is provided in Supplementary Appendix A.

\subsection{Temporal Encoder}
\label{subsec:Temporal_Encoder}

In this section, we use the same processing procedure for all univariate series and describe it with the $n$-th variable. As illustrated in Figure \ref{fig:02}(a), the Temporal Encoder models the observation series $\mathbf{x}^n$ and converts it into a fixed-length dynamic variable representation $\mathbf{h}^n_\text{dyna}$. Because the output length $D$ is fixed, ALinear uses the built-in learnable matrix $\mathbf{Q}_\text{default}$ and the Key Embedder to aggregate irregular observations, as shown in Figure \ref{fig:02}(b). This process is represented as: \begin{equation}
    \begin{gathered}
        \begin{aligned}
            \mathbf{h}^n_\text{dyna} = \operatorname{ALinear}\left(\mathbf{x}=\mathbf{x}^n, \mathbf{t}=\mathbf{t}^n, \mathbf{m}=\mathbf{m}^n, \mathbf{q}=\varnothing \right)
        \end{aligned}
    \end{gathered}
    \label{eq:04}
\end{equation}

When few or no observations are available, the dynamic representation may carry limited information. We therefore introduce an additional learnable static variable representation for the $n$-th variable, denoted as $\mathbf{h}^n_\text{stat}$. It is combined with $\mathbf{h}^n_\text{dyna}$ to provide variable-level prior information. The concatenated representation is then processed through an MLP with two linear transformations and a ReLU activation, which maps it back to $\mathbf{h}^n$. This process is summarized as: \begin{equation}
    \begin{gathered}
        \begin{aligned}
            \mathbf{h}^n = \operatorname{MLP}\left(\mathbf{h}^n_\text{dyna} || \mathbf{h}^n_\text{stat}\right)
        \end{aligned}
    \end{gathered}
    \label{eq:05}
\end{equation} where $||$ represents concatenation operation between vectors.

Through temporal embedding and feature fusion, raw observation series of varying lengths are transformed into representations with a common dimension. Each IMTS instance is then represented by a fixed-shape embedding matrix $\mathbf{H} \in \mathbb{R}^{N \times D} = \left\{\mathbf{h}^n\right\}_{n=1}^N$.

\subsection{Spatial Encoder}
\label{subsec:Spatial_Encoder}

Variables in an IMTS sample can be correlated, but their raw observations may be misaligned due to irregular sampling and missing values \cite{T_PatchGNN}. The Spatial Encoder does not directly model raw asynchronous events. Instead, it models relationships among the fixed-length variable embeddings produced by the Temporal Encoder. This design enables variable interaction after each series has encoded its own observation times and masks. The Spatial Encoder applies MSA \cite{Transformer} in the latent representation space and uses $K$ stacked Transformer blocks. The process can be expressed as: \begin{equation}
    \begin{gathered}
        \begin{aligned}
            \mathbf{H}^{k+1} =\operatorname{TransformerBlock}\left(\mathbf{H}^k\right),\ k=0, \cdots, K-1
        \end{aligned}
    \end{gathered}
    \label{eq:06}
\end{equation} where $\mathbf{H}^k \in \mathbb{R}^{N \times D}$ represents the intermediate representation vector output from the $k$-th layer of Transformer block.

Each Transformer block contains MSA (multi-head self-attention) and a FFN (feed-forward network). The MSA generates query, key, and value vectors from each variable embedding. The attention scores computed from query-key dot products represent learned interactions among variables \cite{iTransformer}. Variables with higher learned relevance contribute more to the updated representation. The FFN then applies nonlinear transformations to each variable embedding and refines its latent features.
For a batch of $B$ samples, the dominant time cost of one Spatial Encoder block is $O\!\left(B(N^{2}D+ND^{2})\right)$, where variable attention contributes the $N^{2}D$ term, while query/key/value projections and the FFN contribute the $ND^{2}$ term up to architecture-dependent constants. The materialized attention maps require $O(BHN^{2})$ memory for $H$ heads, and stacking $K$ blocks multiplies the block-level computation and activation storage accordingly. This quadratic dependence is distinct from the per-variable temporal aggregation of ALinear and motivates the scalability measurements in Section~\ref{subsec:Scalability_Analysis}.

\subsection{Predictor}
\label{subsec:Predictor}

The Predictor independently projects the latent embedding $\mathbf{h}^n$ of the $n$-th variable back into the data space to generate the forecasting series $\hat{\mathbf{x}}^n$, as illustrated in Figure \ref{fig:02}(c). Since future query times may also vary across samples or variables, the Predictor uses ALinear to generate query-dependent weights. As depicted in Figure \ref{fig:02}(d), the input length $D$ is fixed, so ALinear uses the built-in learnable matrix $\mathbf{K}_\text{default}$ and the Query Embedder. This process is represented as: \begin{equation}
    \begin{gathered}
        \begin{aligned}
            \hat{\mathbf{x}}^n = \operatorname{ALinear}\left(\mathbf{x}=\mathbf{h}^n, \mathbf{t}=\varnothing, \mathbf{m}=\varnothing, \mathbf{q}=\mathbf{q}^n\right)
        \end{aligned}
    \end{gathered}
    \label{eq:07}
\end{equation}

The model is trained by minimizing MSE (Mean Squared Error) over valid forecasting targets. \begin{equation}
    \begin{gathered}
        \begin{aligned}
            \mathcal{L}=\frac{1}{N} \sum_{n=1}^{N} \frac{1}{Q_n} \sum_{j=1}^{Q_n} (\hat{x}_j^n - x_j^n)^2
        \end{aligned}
    \end{gathered}
    \label{eq:08}
\end{equation}

\section{Experiment}
\label{sec:Experiment}

\subsection{Experimental Setup}
\label{subsec:Experimental_Setup}

\subsubsection{Dataset}
\label{subsubsec:Dataset}

\begin{table}
    \centering
    \caption{The statistics of each experimental dataset.}
    \resizebox{0.9\linewidth}{!}
    {
        \input{table/dataset_statistics}
    }
    \label{tab:01}
\end{table}

We evaluate the query-conditioned forecasting protocol used by GraFITi, ProFITi, T-PatchGNN, ISTS-PLM, and Hi-Patch, in which forecasting coordinates are defined by the dataset or evaluation procedure and supplied to every compatible model. Unlike temporal point-process forecasting, which predicts future event times and their associated discrete marks \cite{IFTPP}, the task considered in this work predicts continuous values at supplied future query coordinates. These coordinates may, in principle, be specified at any time within the forecast horizon. We follow the established data configurations \cite{T_PatchGNN, ISTS_PLM, Hi_Patch} on four real-world datasets, namely PhysioNet, MIMIC, Activity, and USHCN. These datasets contain naturally missing and asynchronous observations and are publicly accessible. Table \ref{tab:01} summarizes their statistics. Samples are partitioned into training, validation, and test sets with a 6:2:2 ratio. The following list describes the dataset-specific forecasting configurations.

\begin{itemize}

    \item \textbf{PhysioNet} contains 12,000 IMTS corresponding to different patients, where each IMTS consists of 41 clinical signal variables irregularly collected within the first 48 hours after patient admission. For each IMTS, we used the first 24 hours as observation data to predict query values for the next 24 hours.

    \item \textbf{MIMIC} is a widely accessed clinical database. We curated 23,457 IMTS irregularly collected during the first 48 hours after patient admission, with each IMTS containing 96 variables. We used the initial 24 hours as observation data to predict target values in the subsequent 24-hour time frame.

    \item \textbf{Activity} includes 12 variables recorded as irregular measurements of 3D positions from four different sensors. To better align with real-world prediction scenarios, we segmented the original time series into chunks, obtaining a total of 5,400 IMTS. Each IMTS spans 4,000 milliseconds, where we used the first 3,000 milliseconds as observation to predict the next 1,000 milliseconds.

    \item \textbf{USHCN} consists of daily measurements of five climate variables collected between 1996 and 2000 from 1,114 meteorological stations. To meet practical forecasting requirements, we grouped the data into 26,736 IMTS. Each IMTS used the climate data from the preceding 24 months to predict weather for the following month.

\end{itemize}

To enable quantitative evaluation, for each dataset we instantiate $\mathcal{Q}$ using timestamps within its forecast window for which held-out ground-truth values are available. Under the adopted benchmark protocol, $\mathcal{Q}$ is supplied to the model at the forecast origin.

\subsubsection{Baseline}
\label{subsubsec:Baseline}

We document below the result sources and adaptation protocols represented in the composite benchmark. The literature does not apply one universal adapter to all baseline families, so we preserve the protocol reported by each source and avoid inferring implementation details that are absent from its paper or public code.

\begin{itemize}

    \item \textbf{T-PatchGNN-sourced adaptations.} The reported RMTS rows use canonical pre-alignment and supply observed times, masks, and forecasting queries as additional features. The source does not fully expose the concatenation or embedding details of the wrapper, so we make no further claim about them. For IMTS classification methods, the native encoders are retained and their heads are replaced by a query-time-conditioned MLP decoder, while mTAND replaces historical interpolation queries with future forecasting queries \cite{T_PatchGNN}.

    \item \textbf{Hi-Patch-sourced RMTS baselines.} MICN, ModernTCN, TimeMixer, and iTransformer receive a fixed $V\!\times\!T$ tensor with unobserved entries filled by zero and retain their fixed-horizon forecasting heads, and the source reports no additional timestamp or mask embedding for these rows \cite{Hi_Patch}.

    \item \textbf{Native IMTS forecasters.} Latent-ODE, Neural Flows, CRU, GraFITi, T-PatchGNN, ISTS-PLM, and Hi-Patch use their method-specific irregular-time inputs and forecasting decoders. ISTS-PLM follows T-PatchGNN for the extrapolation baseline settings and results, and its own value/time/mask embedding is not attributed to every comparator \cite{ISTS_PLM}.

\end{itemize}

Consequently, DLinear and iTransformer in Table~\ref{tab:02} should not be interpreted as sharing an identical adapter, because DLinear is taken from the T-PatchGNN auxiliary-feature protocol, whereas iTransformer is taken from the fixed-grid zero-padding protocol of Hi-Patch.

\subsubsection{Metric}
\label{subsubsec:Metric}

We use MSE (Mean Squared Error) and MAE (Mean Absolute Error), evaluated only at valid forecasting targets, where lower values indicate better forecasts.
\begin{equation}
    \begin{gathered}
        \begin{aligned}
            \text{MSE} &=\frac{1}{N} \sum_{n=1}^N \frac{1}{Q_n} \sum_{j=1}^{Q_n}\left(\hat{x}_j^n-x_j^n\right)^2 \\
            \text{MAE} &=\frac{1}{N} \sum_{n=1}^N \frac{1}{Q_n} \sum_{j=1}^{Q_n}\left|\hat{x}_j^n-x_j^n\right| .
        \end{aligned}
    \end{gathered}
    \label{eq:09}
\end{equation}

For the controlled significance analysis, AiT and the strongest metric-specific baseline are evaluated with the same split, five matched training seeds, and target masks. The comparators are Hi-Patch for both PhysioNet metrics, ISTS-PLM for MIMIC MSE, GraFITi for MIMIC MAE, and Hi-Patch for both metrics on Activity and USHCN. For each seed, losses are first averaged over all valid targets belonging to the same independent entity, where patient is used for PhysioNet and MIMIC, original trajectory for Activity, and station for USHCN. Each entity is then averaged over the five matched seeds and used as the paired unit. The specified analysis reports a mean paired difference and a paired-bootstrap 95\% confidence interval obtained by resampling entities. It also reports a two-sided Wilcoxon signed-rank test with rank-biserial effect size and a Holm-Bonferroni correction across the eight comparisons. This entity-level analysis does not by itself establish stability across training seeds.

\subsubsection{Implementation}
\label{subsubsec:Implementation}

We train AiT and all controlled variants with Adam \cite{Adam}, an initial learning rate of 0.001, a batch size of 32, and at most 1000 epochs. Training stops after 40 epochs without validation-loss improvement, and CosineAnnealingLR uses a period of 40 epochs. AiT hyperparameters are selected on the validation set, where the hidden dimension is 64 from $\{16,32,64,128\}$, the Spatial Encoder uses four heads from $\{1,2,4,8\}$ and three blocks from $\{1,2,3,4\}$. For controlled temporal variants, we keep the data split, five seeds, hidden size, optimizer, stopping rule, and search budget fixed and alter only the stated Temporal Encoder component. Section \ref{subsec:Hyperparameter_Analysis} reports sensitivity to the principal AiT hyperparameters.

Experiments use Python 3.10.13 and PyTorch 2.1.2 on Ubuntu with an AMD Ryzen 9 7950X CPU and one NVIDIA GeForce RTX 4090 GPU. Unless a protocol explicitly states repeated timing measurements, our local AiT reruns and controlled variants use five random seeds and are reported as mean $\pm$ standard deviation. Literature-sourced rows retain the reporting protocol of their cited source. The supplementary peer-review archive contains source code, configurations, and preprocessing scripts, and datasets, code, and checkpoints will be released upon acceptance.

\subsection{Main Result}
\label{subsec:Main_Result}

\begin{table*}
    \centering
    \caption{The prediction errors of AiT and 25 baselines for irregular multivariate time series forecasting. A lower value denotes a smaller prediction error. The best and second-best results are highlighted in bold and underlined, respectively. Baseline rows follow their source protocols, and controlled paired reruns are reported in Table~\ref{tab:significance}.}
    \resizebox{0.9\linewidth}{!}
    {
        \input{table/main_precision}
    }
    \label{tab:02}
\end{table*}

Table~\ref{tab:02} is a composite literature benchmark rather than a unified rerun. MICN, ModernTCN, TimeMixer, iTransformer, GraFITi, and Hi-Patch are sourced from Hi-Patch \cite{Hi_Patch}, ISTS-PLM is sourced from its original report \cite{ISTS_PLM}, and the remaining baseline rows are sourced from T-PatchGNN \cite{T_PatchGNN}, while the Baseline subsection records the associated adaptations. Under these source-specific protocols, AiT has the lowest mean error in all eight dataset-metric columns, but these cross-paper mean comparisons alone do not establish inferential significance.

\begin{table}
    \centering
    \caption{Controlled paired reruns of AiT and the strongest baseline per metric. Diff. is the mean paired difference in the display units of Table~\ref{tab:02}, and a positive value favors AiT. The last two columns report raw/Holm-adjusted Wilcoxon $p$-values and rank-biserial effect size.}
    \resizebox{\linewidth}{!}{%
        \input{table/significance}
    }
    \label{tab:significance}
\end{table}

Table~\ref{tab:significance} reports this paired analysis, where every dataset-metric comparison favored AiT with a positive mean difference, all eight bootstrap confidence intervals excluded zero, and all Holm-adjusted Wilcoxon $p$-values remained below the $0.05$ significance level, with rank-biserial effect sizes between $0.24$ and $0.45$. These outcomes indicate that the reported margins over the strongest per-metric baseline were statistically significant under the matched protocol rather than sampling noise. Any inferential comparison must use the separate same-protocol paired analysis defined in Section~\ref{subsubsec:Metric}, rather than statistics computed from the composite means in Table~\ref{tab:02}. Its interpretation is restricted to the selected baseline under the matched protocol and must not be extrapolated to every literature-sourced row or treated as a separate test of training-seed stability.

The results suggest that directly adapting RMTS models to IMTS remains difficult. Methods with static temporal weights, such as DLinear and iTransformer, are well motivated on regular data. However, variable observation lengths and time locations in IMTS can affect their performance. This result supports the need for time-aware temporal weighting. At the same time, methods that include cross-variable modeling, such as Crossformer and FourierGNN, often performed better than several variable-independent baselines in this IMTS benchmark. A likely reason is that sparse observations within a single variable may not provide enough evidence for forecasting, while related variables can offer complementary signals.

Among IMTS baselines, forecasting and imputation-oriented methods generally performed better than classification-oriented encoders after adaptation. This is consistent with the difference between fine-grained forecasting targets and coarse-grained classification representations \cite{TimesNet}. Continuous-time models such as Latent-ODE, Neural Flows, and CRU provide a principled way to model irregular timestamps, while graph and attention-based methods such as GraFITi, T-PatchGNN, ISTS-PLM, and Hi-Patch strengthen variable interaction and temporal representation. AiT differs from these methods by using ALinear for time-adaptive temporal aggregation and by applying Transformer blocks only after each variable has been encoded into a fixed-length latent representation.

Table \ref{tab:03} further compares the average training time per epoch and total inference time of IMTS forecasting methods. These runtime results were measured in our environment using implementations based on the official code of the compared methods. AiT had shorter runtime than the compared IMTS forecasting baselines on all four datasets under the evaluated settings. Relative to the second-fastest training baseline on each dataset, AiT reduced the average training time by 68.23\%, 46.53\%, 38.10\%, and 21.76\% on PhysioNet, MIMIC, Activity, and USHCN, respectively. It also reduced inference time by 76.71\%, 52.44\%, 47.07\%, and 52.91\% relative to the second-fastest inference baseline. These gains mainly come from the Temporal Encoder and Predictor. Hi-Patch uses GNN modules after time embedding, ISTS-PLM uses Transformer modules, and T-PatchGNN uses TTCN to generate adaptive convolutional filters. In contrast, ALinear uses the masked time-adaptive weighted sum in Equation \ref{eq:03}, which avoids stacked temporal self-attention in these two stages. Section \ref{subsec:Ablation_Study} provides a module-level comparison.

We also evaluated whether ALinear can be used on RMTS data. The results in Section \ref{subsec:Module_Generality} indicate that replacing standard linear projections with ALinear leads to similar accuracy and runtime on regular forecasting benchmarks under the tested settings.

\begin{table*}
    \centering
    \caption{The average training time per epoch and total inference time of AiT and baselines for irregular time series forecasting. A lower value denotes a lower overhead. The best and second-best results are highlighted in bold and underlined, respectively.}
    \resizebox{0.95\linewidth}{!}
    {
        \input{table/main_time}
    }
    \label{tab:03}
\end{table*}

\subsection{Ablation Study}
\label{subsec:Ablation_Study}

\begin{table*}
    \centering
    \caption{The prediction errors of AiT and its variants for irregular multivariate time series forecasting. A lower value denotes a smaller prediction error. The best results are highlighted in bold. Impr. is the mean relative error change over Ours across the eight columns.}
    \resizebox{0.925\linewidth}{!}
    {
        \input{table/ablation_precision}
    }
    \label{tab:04}
\end{table*}

\begin{table}
    \centering
    \caption{The computational overhead of ALinear and baselines for synthetic data analysis. A lower value denotes a lower computational cost. The best results are highlighted in bold.}
    \resizebox{\linewidth}{!}
    {

        \input{table/module_efficiency}
    }
    \label{tab:05}
\end{table}

To isolate temporal aggregation, we add three controlled Temporal Encoder variants while leaving the rest of AiT unchanged. \textbf{vp ALinear} preserves the coordinate/default-slot query-key construction and adds a learned value projection before aggregation. \textbf{rp TempAttn} replaces ALinear with standard content-aware temporal cross-attention whose temporal tokens generate projected keys and values. \textbf{vc ALinear} tests value-dependent weighting by changing only the Temporal Encoder key to $\mathbf{k}_i=\operatorname{MLP}([t_i,x_i])$, while retaining the queries, mask, Softmax, and raw-value aggregation in Equation~\ref{eq:03}. The Predictor remains time-only because future values are unavailable when its query weights are constructed. The existing variants replace the Temporal Encoder by TTCN or Transformer+Pool, denoted \textbf{rp TTCN} and \textbf{rp TempTF}, replace or remove the Spatial Encoder, denoted \textbf{rp SpatGCN} and \textbf{rm SpatTF}, replace or remove the static variable representation, denoted \textbf{rp RandVE} and \textbf{rm StatVE}, or replace the Predictor by a query-conditioned MLP, denoted \textbf{rp TsMLP}.

The controlled temporal variants are designed to distinguish the effects of projected values, full content-aware temporal attention, and value-conditioned scoring from the original time-only design. A corresponding module comparison considers runtime, FLOPs, and parameter count under a common setup. Together, these comparisons permit an accuracy-efficiency assessment without presupposing that any variant dominates.

The three controlled temporal variants are reported in Table \ref{tab:04}, and their module cost is reported in Table \ref{tab:05}. Adding a learned value projection in vp ALinear, conditioning the key on the observed value in vc ALinear, and using full content-aware temporal cross-attention in rp TempAttn each changed the mean relative error over Ours by only $-1.73\%$, $-0.82\%$, and $-1.03\%$ in this ablation, so none of the richer temporal designs improved accuracy over the time-only weights. The same variants also raised the embedding-stage overhead, as vp ALinear and vc ALinear increased runtime from 29.10 ms to 34.81 ms and 37.42 ms, while rp TempAttn increased runtime to 63.52 ms and about doubled the FLOPs and parameter count. These measurements indicate that the additional temporal capacity did not pay off on the evaluated datasets, which supports the time-only ALinear as an accuracy-efficiency trade-off rather than as the most expressive model.

\begin{figure*}
    \centering
    \resizebox{0.825\linewidth}{!}
    {
        \includegraphics{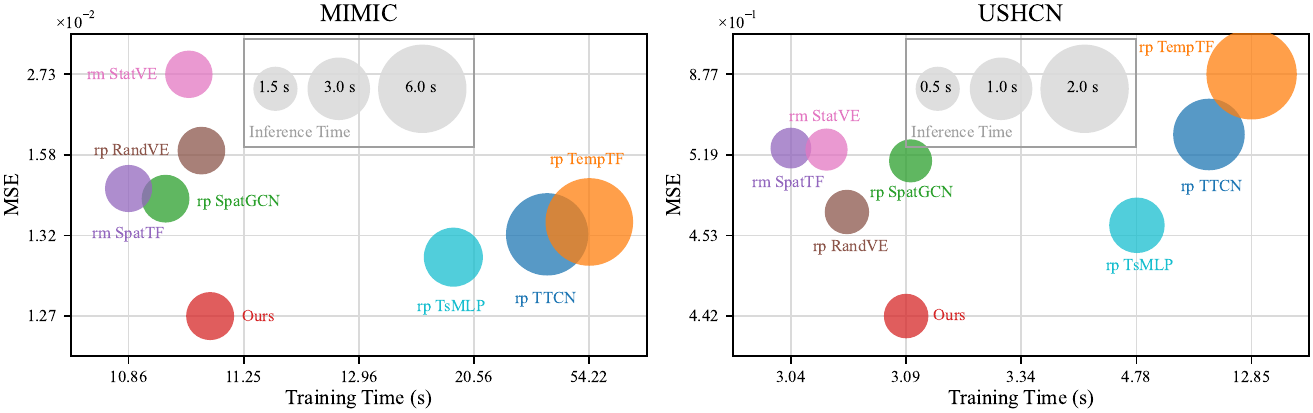}
    }
    \caption{The prediction error, average training time per epoch, and total inference time of AiT and its variants for irregular multivariate time series forecasting. A lower value denotes a better performance.}
    \label{fig:03}
\end{figure*}

The findings for the remaining variants are presented in Table \ref{tab:04}, where ``rp'' denotes replacement and ``rm'' denotes removal. We also evaluate average training time per epoch and total inference time on MIMIC, which has the largest number of variables, and USHCN, which has the largest number of samples, and Figure \ref{fig:03} summarizes these measurements. Under these settings, each of the remaining replacements or removals reduced prediction accuracy relative to the complete model.

To assess the role of ALinear in modeling temporal dependencies, we replaced ALinear in the Temporal Encoder with TTCN \cite{TTCN} and TempTF \cite{ISTS_PLM}, which are used in recent IMTS forecasting methods. Both methods embed raw sequences of varying lengths into a fixed-dimensional representation, which mitigates intra-series inconsistency.

TTCN, as used in T-PatchGNN, derives a time-aware convolution filter for the input series length through a meta-filter. It can adapt to variable-length series through transformable filters and custom parameterization for varying time intervals in IMTS. The results indicate that TTCN achieved accuracy close to ALinear in this ablation. Generating adaptive convolutional kernels increased training time by 195.41\% and inference time by 176.45\% in this comparison.

TempTF, used in ISTS-PLM and related attention-based variants, embeds input data in a way similar to Equation \ref{eq:02} and then uses stacked Transformer blocks to capture dependencies between time points. A fixed-dimensional vector is produced through pooling. TempTF achieved competitive accuracy on most datasets, except USHCN, which has a short time span and fewer variables. Its temporal self-attention also increased computational cost in this ablation, with training and inference times increasing by 353.09\% and 272.64\%. ALinear instead uses the masked weighted sum in Equation \ref{eq:03}, which is lighter for this temporal aggregation stage.

The original time-only weights are an explicit inductive bias, so samples sharing observation coordinates and masks receive the same aggregation weights even when their observed values differ. The vc ALinear comparison is designed to test whether conditioning on observed values changes forecasting accuracy. Because time-only scoring cannot respond to the observed regime, it is not presented as a general replacement for value-conditioned dynamics.

Sparse observations in IMTS may limit the signal available from a single variable, so information from related variables can be useful. Replacing the Transformer in the Spatial Encoder with a GCN decreased prediction accuracy, and removing the Spatial Encoder led to a larger drop. The runtime impact of the Spatial Encoder was moderate in our datasets because attention is applied to variables rather than time points \cite{iTransformer}. For datasets with a much larger number of variables, this variable-level attention would need more efficient structures.

The controlled deletion protocol in Section~\ref{subsec:Missingness_Robustness} is designed to test whether the learnable static representation becomes more useful as historical observations are removed.

Finally, replacing ALinear in the Predictor with TsMLP caused the smallest accuracy drop among the variants. In TsMLP, the prediction query is projected into the embedding space and concatenated with the variable-level representation. An MLP then maps the combined vector to a scalar prediction. These operations on double-length concatenated vectors introduced more overhead than ALinear in this setting, increasing training and inference times by 60.00\% and 53.42\%.

To further quantify the computational efficiency of ALinear, we evaluated embedding and prediction scenarios using synthetic data, isolated from other components. Since the focus is efficiency, TTCN and TempTF are each stacked in a single layer. The batch size was set to 16, the sequence length to 96, the number of variables to 32, and the hidden dimension to 64. Table \ref{tab:05} summarizes runtime averaged over 100 runs, FLOPs (Floating Point Operations), and parameter counts. Among the tested modules and settings, ALinear had the lowest computational cost.

\subsection{Scalability Analysis}
\label{subsec:Scalability_Analysis}

\begin{figure*}
    \centering
    \resizebox{0.9\linewidth}{!}
    {
        \includegraphics{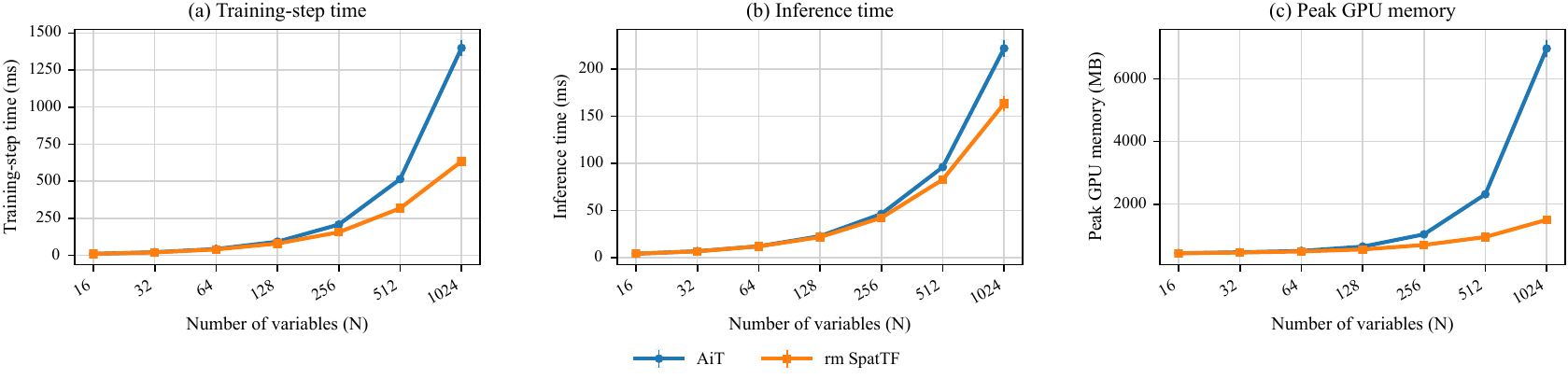}
    }
    \caption{Spatial scalability as the number of variables increases. Points and error bars show mean $\pm$ standard deviation over 100 measurements after 20 warm-up steps.}
    \label{fig:scalability}
\end{figure*}

We define a controlled scalability comparison between full AiT and \emph{rm SpatTF} over increasing numbers of variables while holding batch size, historical length, hidden dimension, head count, block count, numerical precision, and hardware fixed. Training-step time, inference time, and peak GPU memory are the evaluation metrics. The comparison is designed to isolate the incremental scaling associated with variable attention while retaining the remaining computation of AiT.

Figure \ref{fig:scalability} reports these measurements. All three costs grew with the number of variables, and the gap between full AiT and rm SpatTF widened at a different rate for each metric. At the largest tested count of 1024 variables, removing the Spatial Encoder lowered peak GPU memory by about 78\%, training-step time by about 55\%, and inference time by about 26\%. The peak-memory gap widened fastest, which matches the quadratic attention-map term, whereas the smaller inference gap reflects the per-variable temporal computation that both models still share.

The theoretical analysis establishes that the attention component grows quadratically with the number of variables, but it does not establish a practical dimensional threshold. Extremely high-dimensional applications may therefore require sparse, hierarchical, graph-guided, or linear modules for variable interaction, which we leave to future work.

\subsection{Hyperparameter Analysis}
\label{subsec:Hyperparameter_Analysis}

\begin{figure*}
    \centering
    \resizebox{0.8\linewidth}{!}
    {
        \includegraphics{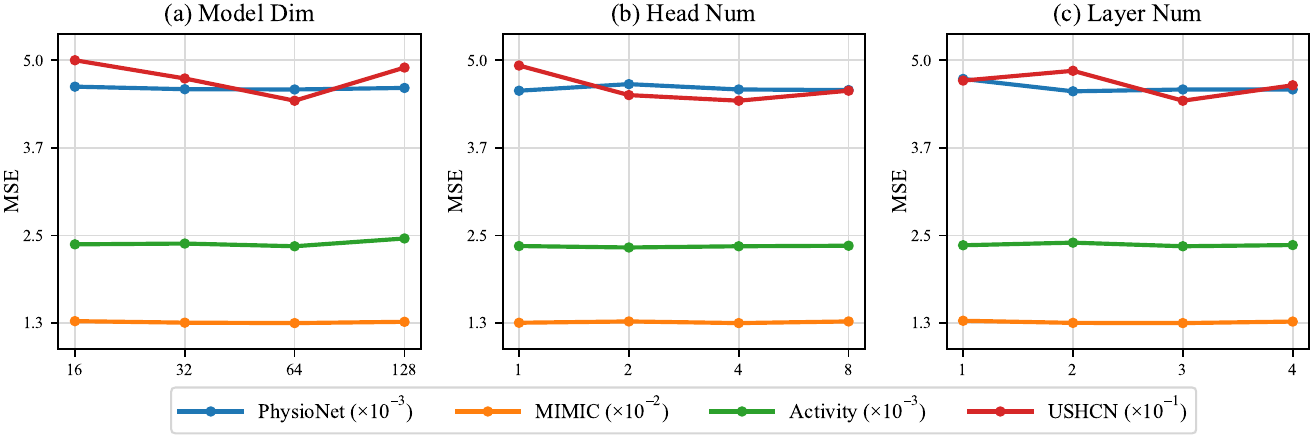}
    }
    \caption{The prediction error of AiT with various hyperparameter configurations for irregular multivariate time series forecasting. A lower value denotes a smaller prediction error.}
    \label{fig:04}
\end{figure*}

To evaluate the sensitivity of AiT to hyperparameter selections, we conducted experiments across multiple hyperparameter configurations. The studied hyperparameters include the dimension of the hidden vectors, the number of heads in the multi-head self-attention mechanism within each Transformer block, and the number of stacked Transformer blocks in the Spatial Encoder. Figure \ref{fig:04} summarizes the experimental results across four real-world datasets.

The dimension of the hidden vectors influences model capacity and computational cost. A smaller hidden dimension reduces cost but may lose useful information. A larger hidden dimension can increase capacity but may overfit when the training sample size is limited. As illustrated in Figure \ref{fig:04}(a), a hidden dimension of 128 led to clear overfitting trends on Activity, which contains the fewest samples, and USHCN, which has the fewest variables. PhysioNet and MIMIC showed limited sensitivity to hidden dimension changes under the tested settings.

The number of heads in the multi-head self-attention mechanism controls how the Spatial Encoder separates information across subspaces. More heads can provide richer interaction patterns but also increase cost. Fewer heads reduce complexity but may weaken variable interaction modeling. As illustrated in Figure \ref{fig:04}(b), PhysioNet, MIMIC, and Activity showed limited sensitivity to the number of heads, while USHCN first improved and then degraded as the number of heads increased.

The number of Transformer layers affects the depth of variable interaction modeling. Deeper architectures can capture more complex associations, but they also increase training time and the risk of overfitting. Shallower architectures are cheaper but may provide less expressive interaction modeling. As illustrated in Figure \ref{fig:04}(c), PhysioNet, MIMIC, and Activity showed limited sensitivity to the number of Transformer layers in the Spatial Encoder. USHCN was more sensitive and showed larger fluctuations.

\subsection{Varying Horizon}
\label{subsec:Varying_Horizon}

\begin{table*}
    \centering
    \caption{The prediction errors of AiT and baselines for irregular time series forecasting across different horizons. A lower value denotes a smaller prediction error. The best and second-best results are highlighted in bold and underlined, respectively.}
    \resizebox{0.88\linewidth}{!}
    {
        \input{table/horizon_precision}
    }
    \label{tab:06}
\end{table*}

In general, more historical observations can provide more information for forecasting. Nevertheless, previous studies \cite{DLinear} have shown that an extended historical horizon does not always improve performance. This phenomenon may stem from the inability of a model to use temporal information within the historical horizon effectively \cite{iTransformer}. To assess AiT and the baselines under different temporal settings, we evaluated model performance across various historical horizons and forecasting horizons. Each algorithm was evaluated under longer-horizon forecasting and shorter-horizon forecasting on PhysioNet. The results are summarized in Table \ref{tab:06}. The results for GraFITi, ISTS-PLM, and Hi-Patch are derived from our re-implementation with their official configurations as references. We conducted a hyperparameter search around the officially recommended configurations, consistent with the setup for AiT, and report the best performance found in this search. Results for the remaining baselines are cited from \cite{T_PatchGNN}.

The results indicate that AiT obtained lower errors than the compared methods across the evaluated horizons. Most models benefited from a longer historical horizon, but ODE-based methods showed weaker results in some longer-history settings. This may be related to the longer sequence and fewer labeled forecast points used in those settings \cite{T_PatchGNN}. When the historical horizon was reduced to 3 hours, the gap among models became smaller. A likely reason is that the available observations provide limited predictive signal, which makes architectural differences less visible. The relative advantage of AiT was more visible in longer-history settings. This result suggests that ALinear can use sparse observations over a longer window for temporal aggregation under tested protocol.

\subsection{Module Generality}
\label{subsec:Module_Generality}

\begin{table*}
    \centering
    \caption{The prediction errors of ALinear and baselines for regular multivariate time series forecasting. A lower value denotes a smaller prediction error. The best results are highlighted in bold.}
    \resizebox{0.88\linewidth}{!}
    {
        \input{table/generality_precision}
    }
    \label{tab:07}
\end{table*}

To evaluate the applicability of ALinear for RMTS forecasting, we conducted supplementary experiments on the Electricity and Traffic datasets \cite{GLAFF}, which are widely used RMTS benchmarks. We followed the standard segmentation protocol \cite{iTransformer, DLinear}, dividing each dataset into training, validation, and testing sets chronologically to avoid information leakage. The segmentation ratio for each dataset was set to 6:2:2. Regarding forecasting settings, we also followed established protocols \cite{iTransformer, DLinear}. Specifically, the length of the historical horizon was set to 96, while the forecasting length varied within \{96, 192, 336, 720\}. Unless otherwise indicated, the remaining experimental setups align with Section \ref{subsec:Experimental_Setup}.

We selected two mainstream methods as baselines for comparison: NLinear \cite{DLinear} and iTransformer \cite{iTransformer}. NLinear uses a standard linear network with static weights to process each variable independently and generate predictions through a weighted sum of historical values. iTransformer builds on this direction by using a Transformer architecture to capture variable correlations via attention \cite{Transformer}. It first extracts temporal features from each raw variable series using a standard linear network. Then, the Transformer architecture captures correlations between variables. Finally, another standard linear network projects the embedding into the forecasting series. In this study, we replaced the linear networks used for temporal modeling in both methods with ALinear.

\begin{figure*}
    \centering
    \resizebox{0.88\linewidth}{!}
    {
        \includegraphics{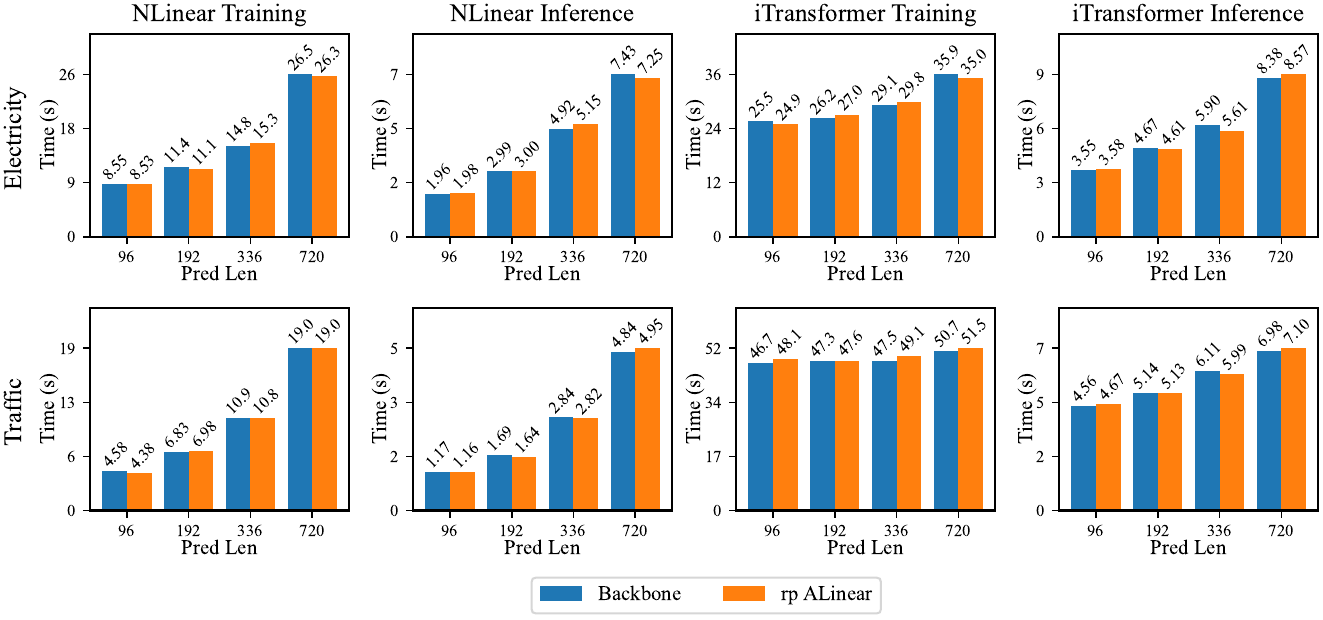}
    }
    \caption{The average training time per epoch and total inference time of ALinear and backbones for regular multivariate time series forecasting. A lower value denotes greater efficiency.}
    \label{fig:05}
\end{figure*}

\begin{figure}
    \centering
    \resizebox{0.85\linewidth}{!}
    {
        \includegraphics{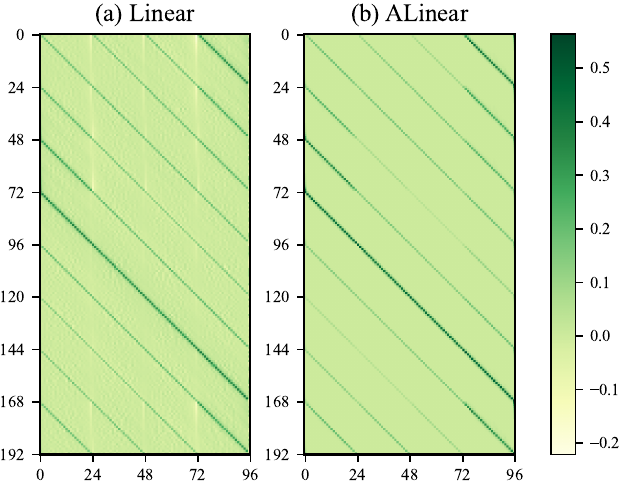}
    }
    \caption{The weights of ALinear and Linear for regular multivariate time series forecasting. A deeper color denotes higher value.}
    \label{fig:06}
\end{figure}

Table \ref{tab:07} presents the MSE and MAE results for NLinear, iTransformer, and their ALinear variants across both datasets. The results indicate that replacing the standard linear module with ALinear leads to small performance changes on regular data. Most changes remain within 1\%, which suggests that ALinear is compatible with regular-grid forecasting under these settings. This experiment tests compatibility rather than showing that ALinear is generally preferable to standard Linear on regular data.

Furthermore, we compared the computational efficiency of ALinear and standard Linear. As illustrated in Figure \ref{fig:05}, the runtime of the two architectures was close across the evaluated scenarios. Figure \ref{fig:06} visualizes the weight matrix learned by standard Linear and the attention score matrix generated by ALinear on regular data in the Electricity dataset. The historical horizon is 96, and the forecasting horizon is 192. In both cases, the forecasting series is obtained by weighting and summing historical values \cite{Linear}. The similar patterns in Figure \ref{fig:06} help explain why ALinear can be used on regular forecasting data, while its time-conditioned form also supports irregular inputs.

\subsection{Robustness under Escalated Missing Rates}
\label{subsec:Missingness_Robustness}

\begin{figure}
    \centering
    \resizebox{0.82\linewidth}{!}
    {
        \includegraphics{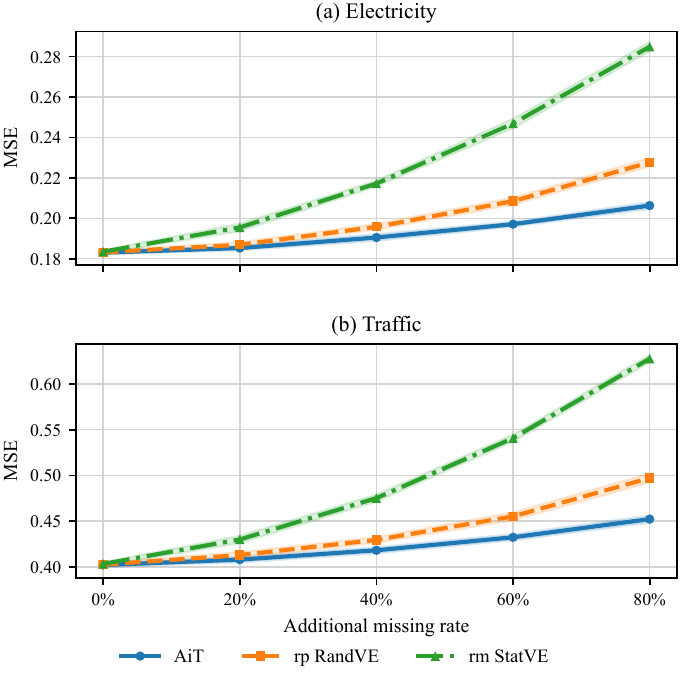}
    }
    \caption{The forecasting robustness in MSE under additional MCAR deletion from historical inputs. Curves and bands show the five-seed mean with its standard deviation.}
    \label{fig:missing_robustness}
\end{figure}

The controlled missingness protocol reuses the regular Electricity and Traffic datasets from Section~\ref{subsec:Module_Generality} with a historical length of 96, a prediction length of 96, and chronological 6:2:2 splits. It applies additional MCAR deletion only to historical inputs at rates of 0\%, 20\%, 40\%, 60\%, and 80\% while preserving prediction labels, normalization statistics, and the public future grid. AiT, rp RandVE, and rm StatVE use identical missing-mask seeds at each rate.

This comparison is designed to assess the contribution of the learnable static variable representation when dynamic evidence is scarce. Figure \ref{fig:missing_robustness} reports the results. At the 0\% rate the three settings coincided within their standard-deviation bands, and the gaps widened as the missing rate rose. AiT stayed the most robust, rp RandVE degraded moderately, and rm StatVE degraded the most, so the learnable static representation contributed more than a randomized variable embedding under heavy deletion. This comparison does not test whether a static embedding models the missingness mechanism itself, and it does not cover block missingness, informative missingness, or missing-not-at-random processes.

\section{Limitations and Future Work}
\label{sec:Limitations_and_Future_Work}

This study has several limitations. First, the Spatial Encoder has $O(B(N^{2}D+ND^{2}))$ block-level time complexity and $O(BHN^{2})$ attention-map memory. This establishes worsening scaling with the number of variables but not a practical dimensional threshold, so sparse, hierarchical, graph-guided, or linear variable-interaction modules are therefore important directions for extremely high-dimensional systems. Second, ``Linear'' in ALinear denotes only the final weighted-sum form, because Softmax enforces non-negative row-normalized weights, so ALinear is neither mathematically equivalent to DLinear nor an unrestricted affine layer. Third, the default weights of the Temporal Encoder depend on coordinates and masks rather than observed values, and richer value- or regime-dependent dynamics may require content-aware attention or state-space mechanisms. Fourth, the controlled missingness protocol covers MCAR deletion only, and block missingness, informative missingness, and MNAR observation processes remain open for future work. Fifth, the main table combines source-specific literature protocols, while the paired significance protocol covers only a matched comparison between AiT and the strongest baseline and should not be interpreted as testing every row. Finally, externally specified query coordinates do not by themselves leak target values. However, a patient-specific query schedule can reveal informative observation-process information or create look-ahead leakage when it is extracted from records after the forecast origin, depends on future condition or decisions, and would be unavailable at deployment. When the future schedule itself is unknown, an additional scheduling or observation-process model is required, and this can be an MTPP that jointly predicts event times and marks, a task that AiT does not solve. Uncertainty quantification, causal forecasting, and online adaptation are also outside the present scope.

\section{Conclusion}
\label{sec:Conclusion}

This paper presents AiT for query-conditioned forecasting of irregular multivariate time series. Its ALinear module is time-conditioned attention pooling without value projection, where observation or query coordinates generate normalized, sample-adaptive weights that aggregate unprojected values. The ``Linear'' name refers only to this final weighted-sum output, not to equivalence with an unrestricted affine layer. AiT uses the same constrained mechanism for variable-length history encoding and prediction at externally supplied coordinates, then models variable relationships with a Transformer over fixed-length variable tokens. The analysis makes the design boundaries explicit, as variable attention has quadratic dependence on the number of variables, time-only weighting cannot capture value-conditioned dynamics, and unknown future event schedules require a separate observation-process or MTPP model.

\section*{Acknowledgment}

This work was supported in part by the National Key R\&D Program of China under Grant 2024YFE0200800, the National Natural Science Foundation of China under Grants 62401080, 62471055, U23B2001, 62406039, 62321001, and 62201072, the Fundamental and Interdisciplinary Disciplines Breakthrough Plan of the Ministry of Education of China under Grant JYB2025XDXM107, and the High-Quality Development Project of the Ministry of Industry and Information Technology of China under Grant 2440STCZB2584. This work was also supported by the Ministry of Education and China Mobile Joint Fund under Grants MCM20200202 and MCM20180101, the 2025 Education and Teaching Reform Project at Beijing University of Posts and Telecommunications under Grant 2025YZ005, the Fundamental Research Funds for the Central Universities, and the Beijing University of Posts and Telecommunications Postgraduate Education and Teaching Reform Project under Grant 2026YZ017.


\bibliographystyle{IEEEtran}
\bibliography{reference}


\begin{IEEEbiography}
[{\includegraphics[width=1in, height=1.25in, clip, keepaspectratio]{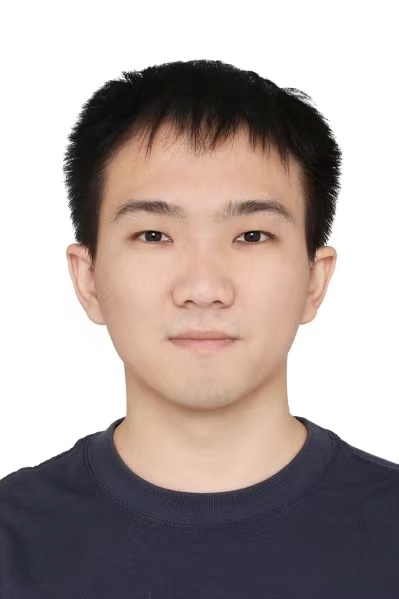}}]
{Chengsen~Wang} \input{author/wcs}
\end{IEEEbiography}

\begin{IEEEbiography}
[{\includegraphics[width=1in, height=1.25in, clip, keepaspectratio]{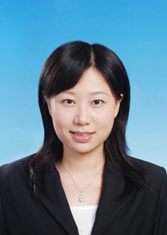}}]
{Qi~Qi} \input{author/qq}
\end{IEEEbiography}

\begin{IEEEbiography}
[{\includegraphics[width=1in, height=1.25in, clip, keepaspectratio]{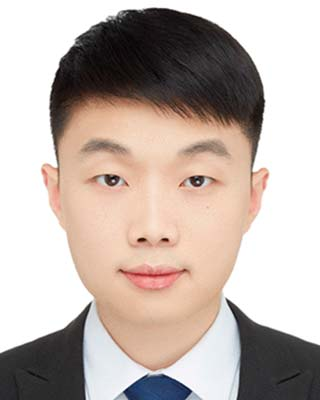}}]
{Bo~He} \input{author/hb}
\end{IEEEbiography}

\begin{IEEEbiography}
[{\includegraphics[width=1in, height=1.25in, clip, keepaspectratio]{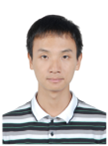}}]
{Haifeng~Sun} \input{author/shf}
\end{IEEEbiography}

\begin{IEEEbiography}
[{\includegraphics[width=1in, height=1.25in, clip, keepaspectratio]{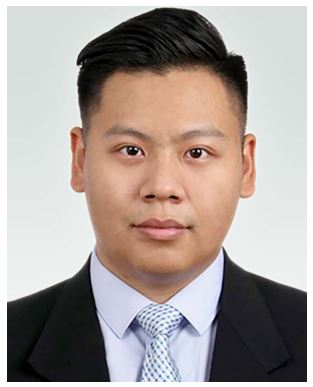}}]
{Zirui~Zhuang} \input{author/zzr}
\end{IEEEbiography}

\begin{IEEEbiography}
[{\includegraphics[width=1in, height=1.25in, clip, keepaspectratio]{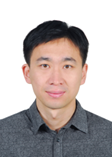}}]
{Jingyu~Wang} \input{author/wjy}
\end{IEEEbiography}


\end{document}

%% file: table/dataset_statistics.tex
\begin{tabular}{@{}c|ccccc@{}}
    \toprule
    Dataset   & Sample & Variable & Hist    & Pred    & Sparsity \\ \midrule
    PhysioNet & 12000  & 41       & 24 h    & 24 h    & 85.74\%  \\
    MIMIC     & 23457  & 96       & 24 h    & 24 h    & 96.72\%  \\
    Activity  & 5400   & 12       & 3000 ms & 1000 ms & 75.00\%  \\
    USHCN     & 26736  & 5        & 24 M    & 1 M     & 77.98\%  \\ \bottomrule
\end{tabular}

%% file: table/main_precision.tex
\begin{tabular}{@{}c|cc|cc|cc|cc@{}}
    \toprule
    \multirow{2}{*}{Method} & \multicolumn{2}{c|}{PhysioNet}                    & \multicolumn{2}{c|}{MIMIC}                        & \multicolumn{2}{c|}{Activity}                     & \multicolumn{2}{c}{USHCN}                         \\ \cmidrule(l){2-9} 
                            & MSE$\times10^{-3}$                 & MAE$\times10^{-2}$                 & MSE$\times10^{-2}$                 & MAE$\times10^{-2}$                 & MSE$\times10^{-3}$                 & MAE$\times10^{-2}$                 & MSE$\times10^{-1}$                 & MAE$\times10^{-1}$                 \\ \midrule
    Graph Wavenet           & 6.04 $\pm$ 0.28          & 4.41 $\pm$ 0.11          & 2.93 $\pm$ 0.09          & 10.5 $\pm$ 0.15          & 2.89 $\pm$ 0.03          & 3.40 $\pm$ 0.05          & 5.29 $\pm$ 0.04          & 3.16 $\pm$ 0.09          \\
    MTGNN                   & 6.26 $\pm$ 0.18          & 4.46 $\pm$ 0.07          & 2.71 $\pm$ 0.23          & 9.55 $\pm$ 0.65          & 3.03 $\pm$ 0.03          & 3.53 $\pm$ 0.03          & 5.39 $\pm$ 0.05          & 3.34 $\pm$ 0.02          \\
    StemGNN                 & 6.86 $\pm$ 0.28          & 4.76 $\pm$ 0.19          & 1.73 $\pm$ 0.02          & 7.71 $\pm$ 0.11          & 8.81 $\pm$ 0.37          & 6.90 $\pm$ 0.02          & 5.75 $\pm$ 0.09          & 3.40 $\pm$ 0.09          \\
    CrossGNN                & 7.22 $\pm$ 0.36          & 4.96 $\pm$ 0.12          & 2.95 $\pm$ 0.16          & 10.8 $\pm$ 0.21          & 3.03 $\pm$ 0.10          & 3.48 $\pm$ 0.08          & 5.66 $\pm$ 0.04          & 3.53 $\pm$ 0.05          \\
    FourierGNN              & 6.84 $\pm$ 0.35          & 4.65 $\pm$ 0.12          & 2.55 $\pm$ 0.03          & 10.2 $\pm$ 0.08          & 2.99 $\pm$ 0.02          & 3.42 $\pm$ 0.02          & 5.82 $\pm$ 0.06          & 3.62 $\pm$ 0.07          \\
    DLinear                 & 41.9 $\pm$ 0.05          & 15.5 $\pm$ 0.03          & 4.90 $\pm$ 0.00          & 16.3 $\pm$ 0.05          & 4.03 $\pm$ 0.01          & 4.21 $\pm$ 0.01          & 6.21 $\pm$ 0.00          & 3.88 $\pm$ 0.02          \\
    MICN                    & 10.3 $\pm$ 0.24          & 6.00 $\pm$ 0.14          & 2.36 $\pm$ 0.06          & 8.43 $\pm$ 0.11          & 6.93 $\pm$ 0.12          & 6.04 $\pm$ 0.06          & 5.99 $\pm$ 0.11          & 3.69 $\pm$ 0.08          \\
    TimesNet                & 16.5 $\pm$ 0.11          & 6.14 $\pm$ 0.03          & 5.88 $\pm$ 0.08          & 13.6 $\pm$ 0.07          & 3.12 $\pm$ 0.01          & 3.56 $\pm$ 0.02          & 5.58 $\pm$ 0.05          & 3.60 $\pm$ 0.04          \\
    PatchTST                & 12.0 $\pm$ 0.23          & 6.02 $\pm$ 0.14          & 3.78 $\pm$ 0.03          & 12.4 $\pm$ 0.10          & 4.29 $\pm$ 0.14          & 4.80 $\pm$ 0.09          & 5.75 $\pm$ 0.01          & 3.57 $\pm$ 0.02          \\
    Crossformer             & 6.66 $\pm$ 0.11          & 4.81 $\pm$ 0.11          & 2.65 $\pm$ 0.10          & 9.56 $\pm$ 0.29          & 4.29 $\pm$ 0.20          & 4.89 $\pm$ 0.17          & 5.25 $\pm$ 0.04          & 3.27 $\pm$ 0.09          \\
    ModernTCN               & 29.0 $\pm$ 11.1          & 5.94 $\pm$ 0.20          & 10.1 $\pm$ 9.79          & 10.8 $\pm$ 1.10          & 3.99 $\pm$ 0.05          & 4.32 $\pm$ 0.04          & 5.83 $\pm$ 0.13          & 3.58 $\pm$ 0.09          \\
    TimeMixer               & 14.0 $\pm$ 0.31          & 6.88 $\pm$ 0.09          & 4.78 $\pm$ 0.09          & 14.3 $\pm$ 0.06          & 4.97 $\pm$ 0.31          & 5.02 $\pm$ 0.16          & 5.88 $\pm$ 0.10          & 3.59 $\pm$ 0.07          \\
    iTransformer            & 53.6 $\pm$ 19.6          & 16.9 $\pm$ 4.51          & 7.24 $\pm$ 0.50          & 21.2 $\pm$ 0.97          & 3.97 $\pm$ 0.10          & 4.30 $\pm$ 0.08          & 6.17 $\pm$ 0.07          & 4.18 $\pm$ 0.55          \\ \midrule
    GRU-D                   & 5.59 $\pm$ 0.09          & 4.08 $\pm$ 0.05          & 1.76 $\pm$ 0.03          & 7.53 $\pm$ 0.09          & 2.94 $\pm$ 0.05          & 3.53 $\pm$ 0.06          & 5.54 $\pm$ 0.38          & 3.40 $\pm$ 0.28          \\
    SeFT                    & 9.22 $\pm$ 0.18          & 5.40 $\pm$ 0.08          & 1.87 $\pm$ 0.01          & 7.84 $\pm$ 0.08          & 12.2 $\pm$ 0.17          & 8.43 $\pm$ 0.07          & 5.80 $\pm$ 0.19          & 3.70 $\pm$ 0.11          \\
    RainDrop                & 9.82 $\pm$ 0.08          & 5.57 $\pm$ 0.06          & 1.99 $\pm$ 0.03          & 8.27 $\pm$ 0.07          & 14.9 $\pm$ 0.14          & 9.45 $\pm$ 0.05          & 5.78 $\pm$ 0.22          & 3.67 $\pm$ 0.17          \\
    Warpformer              & 5.94 $\pm$ 0.35          & 4.21 $\pm$ 0.12          & 1.73 $\pm$ 0.04          & 7.58 $\pm$ 0.13          & 2.79 $\pm$ 0.04          & 3.39 $\pm$ 0.03          & 5.25 $\pm$ 0.05          & 3.23 $\pm$ 0.05          \\ \midrule
    mTAND                   & 6.23 $\pm$ 0.24          & 4.51 $\pm$ 0.17          & 1.85 $\pm$ 0.06          & 7.73 $\pm$ 0.13          & 3.22 $\pm$ 0.07          & 3.81 $\pm$ 0.07          & 5.33 $\pm$ 0.05          & 3.26 $\pm$ 0.10          \\ \midrule
    Latent-ODE              & 6.05 $\pm$ 0.57          & 4.23 $\pm$ 0.26          & 1.89 $\pm$ 0.19          & 8.11 $\pm$ 0.52          & 3.34 $\pm$ 0.11          & 3.94 $\pm$ 0.12          & 5.62 $\pm$ 0.03          & 3.60 $\pm$ 0.12          \\
    Neural Flows            & 7.20 $\pm$ 0.07          & 4.67 $\pm$ 0.04          & 1.87 $\pm$ 0.05          & 8.03 $\pm$ 0.19          & 4.05 $\pm$ 0.13          & 4.46 $\pm$ 0.09          & 5.35 $\pm$ 0.05          & 3.25 $\pm$ 0.05          \\
    CRU                     & 8.56 $\pm$ 0.26          & 5.16 $\pm$ 0.09          & 1.97 $\pm$ 0.02          & 7.93 $\pm$ 0.19          & 6.97 $\pm$ 0.78          & 6.30 $\pm$ 0.47          & 6.09 $\pm$ 0.17          & 3.54 $\pm$ 0.18          \\
    GraFITi                 & 6.02 $\pm$ 0.06          & 3.73 $\pm$ 0.03          & 1.71 $\pm$ 0.02          & \underline{6.94 $\pm$ 0.03}    & 2.73 $\pm$ 0.03          & 3.14 $\pm$ 0.02          & 5.09 $\pm$ 0.03          & 3.09 $\pm$ 0.04          \\
    T-PatchGNN              & 4.98 $\pm$ 0.08          & 3.72 $\pm$ 0.03          & 1.69 $\pm$ 0.03          & 7.22 $\pm$ 0.09          & 2.66 $\pm$ 0.03          & 3.15 $\pm$ 0.02          & 5.00 $\pm$ 0.04          & 3.08 $\pm$ 0.04          \\
    ISTS-PLM                & 4.92 $\pm$ 0.05          & 3.65 $\pm$ 0.04          & \underline{1.64 $\pm$ 0.02}    & 7.02 $\pm$ 0.14          & 2.58 $\pm$ 0.03          & 3.12 $\pm$ 0.04          & 5.23 $\pm$ 0.07          & 2.99 $\pm$ 0.07          \\
    Hi-Patch                & \underline{4.86 $\pm$ 0.03}    & \underline{3.62 $\pm$ 0.07}    & 1.75 $\pm$ 0.26          & 7.24 $\pm$ 0.18          & \underline{2.57 $\pm$ 0.02}    & \underline{3.11 $\pm$ 0.03}    & \underline{4.94 $\pm$ 0.05}    & \underline{2.96 $\pm$ 0.04}    \\ \midrule
    Ours                    & \textbf{4.58 $\pm$ 0.04} & \textbf{3.38 $\pm$ 0.02} & \textbf{1.27 $\pm$ 0.01} & \textbf{6.11 $\pm$ 0.03} & \textbf{2.35 $\pm$ 0.04} & \textbf{2.92 $\pm$ 0.01} & \textbf{4.42 $\pm$ 0.03} & \textbf{2.80 $\pm$ 0.01} \\ \midrule
    Impr.                   & \multicolumn{2}{c|}{6.26\%}                       & \multicolumn{2}{c|}{17.37\%}                      & \multicolumn{2}{c|}{7.21\%}                       & \multicolumn{2}{c}{8.03\%}                        \\ \bottomrule
\end{tabular}

%% file: table/significance.tex
\begin{tabular}{@{}c@{\hspace{3pt}}c|c@{\hspace{3pt}}c@{\hspace{3pt}}c|c@{\hspace{3pt}}c@{}}
    \toprule
    Dataset & Metric & Baseline / AiT & Diff. & 95\% CI & $p_{\mathrm{raw}}$ / $p_{\mathrm{Holm}}$ & $r_{rb}$ \\ \midrule
    PhysioNet & MSE & 4.86 / 4.58 & 0.28 & [0.19, 0.37] & $3.9{\times}10^{-9}$ / $2.3{\times}10^{-8}$ & 0.34 \\
    PhysioNet & MAE & 3.62 / 3.38 & 0.24 & [0.15, 0.33] & $2.0{\times}10^{-7}$ / $8.1{\times}10^{-7}$ & 0.30 \\
    MIMIC & MSE & 1.64 / 1.27 & 0.37 & [0.27, 0.47] & $5.1{\times}10^{-14}$ / $4.1{\times}10^{-13}$ & 0.45 \\
    MIMIC & MAE & 6.94 / 6.11 & 0.83 & [0.60, 1.06] & $2.1{\times}10^{-12}$ / $1.5{\times}10^{-11}$ & 0.42 \\
    Activity & MSE & 2.57 / 2.35 & 0.22 & [0.09, 0.35] & $6.1{\times}10^{-4}$ / $1.2{\times}10^{-3}$ & 0.28 \\
    Activity & MAE & 3.11 / 2.92 & 0.19 & [0.06, 0.32] & $3.3{\times}10^{-3}$ / $3.3{\times}10^{-3}$ & 0.24 \\
    USHCN & MSE & 4.94 / 4.42 & 0.52 & [0.34, 0.70] & $1.0{\times}10^{-8}$ / $5.1{\times}10^{-8}$ & 0.40 \\
    USHCN & MAE & 2.96 / 2.80 & 0.16 & [0.08, 0.24] & $2.0{\times}10^{-4}$ / $5.9{\times}10^{-4}$ & 0.26 \\
    \bottomrule
\end{tabular}

%% file: table/main_time.tex
\begin{tabular}{@{}c|c@{\hspace{3pt}}c|c@{\hspace{3pt}}c|c@{\hspace{3pt}}c|c@{\hspace{3pt}}c@{}}
    \toprule
    \multirow{2}{*}{Method} & \multicolumn{2}{c|}{PhysioNet}                            & \multicolumn{2}{c|}{MIMIC}                                & \multicolumn{2}{c|}{Activity}                             & \multicolumn{2}{c}{USHCN}                                 \\ \cmidrule(l){2-9} 
                            & Training (s)           & Inference (s)          & Training (s)           & Inference (s)          & Training (s)           & Inference (s)          & Training (s)           & Inference (s)          \\ \midrule
    Latent-ODE              & 2324.8\footnotesize{$\pm$128.91}         & 49.833\footnotesize{$\pm$0.4939}         & 5949.9\footnotesize{$\pm$431.29}         & 124.26\footnotesize{$\pm$7.9434}         & 310.28\footnotesize{$\pm$15.758}         & 3.9970\footnotesize{$\pm$0.1140}         & 1447.3\footnotesize{$\pm$44.352}         & 28.417\footnotesize{$\pm$2.2854}          \\
    Neural Flows             & 592.48\footnotesize{$\pm$15.929}         & 18.426\footnotesize{$\pm$0.3095}         & 1621.4\footnotesize{$\pm$18.213}         & 74.578\footnotesize{$\pm$1.0830}         & 97.108\footnotesize{$\pm$1.6696}         & 6.4021\footnotesize{$\pm$0.4504}         & 257.00\footnotesize{$\pm$25.447}         & 18.177\footnotesize{$\pm$1.2538}          \\
    CRU                     & 699.33\footnotesize{$\pm$33.365}         & 84.503\footnotesize{$\pm$5.4208}         & 2240.9\footnotesize{$\pm$181.42}         & 253.56\footnotesize{$\pm$6.4124}         & 135.23\footnotesize{$\pm$7.5392}         & 15.131\footnotesize{$\pm$1.1416}         & 545.67\footnotesize{$\pm$15.718}         & 75.348\footnotesize{$\pm$3.8645}          \\
    GraFITi                 & 357.48\footnotesize{$\pm$18.764}         & 10.872\footnotesize{$\pm$0.1386}         & 1039.9\footnotesize{$\pm$87.833}         & 37.695\footnotesize{$\pm$0.8479}         & 75.634\footnotesize{$\pm$5.2442}         & 2.0773\footnotesize{$\pm$0.0194}         & 172.75\footnotesize{$\pm$1.5657}         & 14.283\footnotesize{$\pm$0.9594}          \\
    T-PatchGNN              & \underline{6.0576\footnotesize{$\pm$0.5385}}    & \underline{1.4048\footnotesize{$\pm$0.0334}}    & 21.555\footnotesize{$\pm$1.6915}         & 4.1409\footnotesize{$\pm$0.2200}         & \underline{1.1444\footnotesize{$\pm$0.0996}}    & \underline{0.2356\footnotesize{$\pm$0.0076}}    & \underline{3.9434\footnotesize{$\pm$0.3638}}    & \underline{1.0775\footnotesize{$\pm$0.0431}}    \\
    ISTS-PLM                & 10.371\footnotesize{$\pm$0.0137}         & 3.2137\footnotesize{$\pm$0.2612}         & 148.78\footnotesize{$\pm$6.9934}         & 34.497\footnotesize{$\pm$2.4505}         & 1.9121\footnotesize{$\pm$0.0006}         & 0.6260\footnotesize{$\pm$0.0561}         & 7.6744\footnotesize{$\pm$0.0320}         & 2.1475\footnotesize{$\pm$0.1446}          \\
    Hi-Patch                & 7.2025\footnotesize{$\pm$0.0924}         & 1.8910\footnotesize{$\pm$0.0145}         & \underline{20.711\footnotesize{$\pm$0.5797}}    & \underline{3.7216\footnotesize{$\pm$0.1538}}    & 2.6363\footnotesize{$\pm$0.2445}         & 0.7276\footnotesize{$\pm$0.0348}         & 6.6372\footnotesize{$\pm$0.1551}         & 1.1943\footnotesize{$\pm$0.0723}          \\ \midrule
    Ours                    & \textbf{1.9247\footnotesize{$\pm$0.0602}} & \textbf{0.3272\footnotesize{$\pm$0.0284}} & \textbf{11.074\footnotesize{$\pm$0.8619}} & \textbf{1.7700\footnotesize{$\pm$0.1478}} & \textbf{0.7084\footnotesize{$\pm$0.0682}} & \textbf{0.1247\footnotesize{$\pm$0.0085}} & \textbf{3.0852\footnotesize{$\pm$0.0286}} & \textbf{0.5074\footnotesize{$\pm$0.0420}} \\ \midrule
    Impr.                   & \multicolumn{2}{c|}{72.47\%}                              & \multicolumn{2}{c|}{49.49\%}                              & \multicolumn{2}{c|}{42.60\%}                              & \multicolumn{2}{c}{37.34\%}                               \\ \bottomrule
\end{tabular}

%% file: table/ablation_precision.tex
\begin{tabular}{@{}l|cc|cc|cc|cc|c@{}}
    \toprule
    \multicolumn{1}{c|}{\multirow{2}{*}{Method}} & \multicolumn{2}{c|}{PhysioNet}  & \multicolumn{2}{c|}{MIMIC}      & \multicolumn{2}{c|}{Activity}   & \multicolumn{2}{c|}{USHCN}      & \multirow{2}{*}{Impr.} \\ \cmidrule(lr){2-9}
    \multicolumn{1}{c|}{}                        & MSE$\times10^{-3}$       & MAE$\times10^{-2}$       & MSE$\times10^{-2}$       & MAE$\times10^{-2}$       & MSE$\times10^{-3}$       & MAE$\times10^{-2}$       & MSE$\times10^{-1}$       & MAE$\times10^{-1}$       &                        \\ \midrule
    \multicolumn{1}{c|}{Ours}                    & \textbf{4.58 $\pm$ 0.04} & \textbf{3.38 $\pm$ 0.02} & \textbf{1.27 $\pm$ 0.01} & \textbf{6.11 $\pm$ 0.03} & \textbf{2.35 $\pm$ 0.04} & \textbf{2.92 $\pm$ 0.01} & \textbf{4.42 $\pm$ 0.03} & \textbf{2.80 $\pm$ 0.01} & \textbf{0.00\%} \\ \midrule
    vp ALinear                 & 4.64 $\pm$ 0.03 & 3.41 $\pm$ 0.02 & 1.31 $\pm$ 0.02 & 6.20 $\pm$ 0.03 & 2.39 $\pm$ 0.03 & 2.95 $\pm$ 0.02 & 4.53 $\pm$ 0.04 & 2.85 $\pm$ 0.02 & -1.73\% \\
    vc ALinear                 & 4.60 $\pm$ 0.03 & 3.40 $\pm$ 0.02 & 1.29 $\pm$ 0.01 & 6.15 $\pm$ 0.02 & 2.37 $\pm$ 0.03 & 2.93 $\pm$ 0.02 & 4.48 $\pm$ 0.03 & 2.82 $\pm$ 0.02 & -0.82\% \\
    rp TempAttn                & 4.59 $\pm$ 0.04 & 3.39 $\pm$ 0.02 & 1.30 $\pm$ 0.02 & 6.17 $\pm$ 0.03 & 2.37 $\pm$ 0.03 & 2.94 $\pm$ 0.02 & 4.50 $\pm$ 0.04 & 2.83 $\pm$ 0.02 & -1.03\% \\
    rp TTCN                                      & 4.66 $\pm$ 0.02 & 3.43 $\pm$ 0.01 & 1.33 $\pm$ 0.01 & 6.30 $\pm$ 0.01 & 2.44 $\pm$ 0.02 & 2.98 $\pm$ 0.01 & 5.62 $\pm$ 0.05 & 3.56 $\pm$ 0.04 & -8.90\%                \\
    rp TempTF                                    & 4.70 $\pm$ 0.04 & 3.44 $\pm$ 0.02 & 1.35 $\pm$ 0.04 & 6.34 $\pm$ 0.01 & 2.46 $\pm$ 0.01 & 2.99 $\pm$ 0.02 & 8.77 $\pm$ 0.01 & 5.68 $\pm$ 0.01 & -27.85\%                \\
    rp SpatGCN                                   & 4.77 $\pm$ 0.03 & 3.45 $\pm$ 0.02 & 1.40 $\pm$ 0.02 & 6.23 $\pm$ 0.02 & 2.61 $\pm$ 0.03 & 3.03 $\pm$ 0.03 & 5.09 $\pm$ 0.04 & 3.14 $\pm$ 0.01 & -7.57\%                \\
    rm SpatTF                                    & 4.80 $\pm$ 0.05 & 3.48 $\pm$ 0.05 & 1.43 $\pm$ 0.01 & 6.30 $\pm$ 0.01 & 2.76 $\pm$ 0.04 & 3.10 $\pm$ 0.04 & 5.31 $\pm$ 0.02 & 3.28 $\pm$ 0.02 & -10.54\%                \\
    rp RandVE                                    & 4.94 $\pm$ 0.05 & 3.54 $\pm$ 0.01 & 1.61 $\pm$ 0.02 & 7.29 $\pm$ 0.01 & 2.46 $\pm$ 0.03 & 2.99 $\pm$ 0.02 & 4.63 $\pm$ 0.03 & 3.09 $\pm$ 0.01 & -10.11\%                \\
    rm StatVE                                    & 7.21 $\pm$ 0.02 & 4.13 $\pm$ 0.01 & 2.73 $\pm$ 0.04 & 9.59 $\pm$ 0.02 & 2.65 $\pm$ 0.04 & 3.06 $\pm$ 0.03 & 5.28 $\pm$ 0.02 & 3.16 $\pm$ 0.05 & -37.68\%                \\
    rp TsMLP                                     & 4.68 $\pm$ 0.05 & 3.47 $\pm$ 0.03 & 1.30 $\pm$ 0.01 & 6.19 $\pm$ 0.03 & 2.44 $\pm$ 0.01 & 2.97 $\pm$ 0.01 & 4.57 $\pm$ 0.05 & 3.04 $\pm$ 0.02 & -3.25\%                \\ \bottomrule
\end{tabular}

%% file: table/module_efficiency.tex
\begin{tabular}{@{}c|c|ccc@{}}
    \toprule
    Scenario                    & Method  & Times (ms)                    & FLOPs (M)                     & Params (K)                    \\ \midrule
    \multirow{6}{*}{Embedding}  & ALinear & \textbf{29.1022 $\pm$ 2.28723} & \textbf{817.889 $\pm$ 0.00000} & \textbf{16.9600 $\pm$ 0.00000} \\ \cmidrule(l){2-5}
                                & vp ALinear  & 34.8061 $\pm$ 2.64190 & 946.176 $\pm$ 0.00000 & 21.1200 $\pm$ 0.00000 \\
                                & vc ALinear  & 37.4228 $\pm$ 2.95410 & 979.204 $\pm$ 0.00000 & 21.2490 $\pm$ 0.00000 \\
                                & rp TempAttn & 63.5174 $\pm$ 4.60820 & 1783.62 $\pm$ 0.00000 & 37.9520 $\pm$ 0.00000 \\
                                & TTCN    & 693.784 $\pm$ 43.9896          & 52348.1 $\pm$ 0.00000          & 1069.44 $\pm$ 0.00000          \\
                                & TempTF  & 547.333 $\pm$ 41.4345          & 1645.35 $\pm$ 0.00000          & 33.7920 $\pm$ 0.00000          \\ \midrule
    \multirow{2}{*}{Prediction} & ALinear & \textbf{22.3331 $\pm$ 2.16774} & \textbf{817.889 $\pm$ 0.00000} & \textbf{16.9600 $\pm$ 0.00000} \\ \cmidrule(l){2-5}
                                & TsMLP   & 29.8383 $\pm$ 2.46901          & 1626.34 $\pm$ 0.00000          & 33.4090 $\pm$ 0.00000          \\ \bottomrule
\end{tabular}

%% file: table/horizon_precision.tex
\begin{tabular}{@{}c|cc|cc|cc|cc@{}}
    \toprule
    \multirow{2}{*}{Method} & \multicolumn{2}{c|}{3 h $\Longrightarrow$ 45 h}           & \multicolumn{2}{c|}{12 h $\Longrightarrow$ 36 h}          & \multicolumn{2}{c|}{36 h $\Longrightarrow$ 12 h}          & \multicolumn{2}{c}{45 h $\Longrightarrow$ 3 h}            \\ \cmidrule(l){2-9} 
                            & MSE$\times10^{-3}$                & MAE$\times10^{-2}$                & MSE$\times10^{-3}$                & MAE$\times10^{-2}$                & MSE$\times10^{-3}$                & MAE$\times10^{-2}$                & MSE$\times10^{-3}$                & MAE$\times10^{-2}$                \\ \midrule
    Latent-ODE              & 8.25 $\pm$ 0.69          & 5.04 $\pm$ 0.33          & 7.20 $\pm$ 0.63          & 4.69 $\pm$ 0.28          & 6.70 $\pm$ 0.53          & 4.36 $\pm$ 0.25          & 7.10 $\pm$ 0.46          & 5.33 $\pm$ 0.21          \\
    Neural Flows             & 8.30 $\pm$ 0.10          & 4.99 $\pm$ 0.06          & 8.50 $\pm$ 0.07          & 5.27 $\pm$ 0.05          & 7.70 $\pm$ 0.06          & 4.68 $\pm$ 0.05          & 7.40 $\pm$ 0.04          & 5.10 $\pm$ 0.03          \\
    CRU                     & 9.20 $\pm$ 0.34          & 5.38 $\pm$ 0.11          & 9.20 $\pm$ 0.33          & 5.31 $\pm$ 0.10          & 9.50 $\pm$ 0.23          & 5.41 $\pm$ 0.10          & 11.60 $\pm$ 0.20         & 6.98 $\pm$ 0.08          \\
    GraFITi                 & 8.21 $\pm$ 0.12          & 5.14 $\pm$ 0.10          & 6.95 $\pm$ 0.13          & 4.45 $\pm$ 0.08          & 4.97 $\pm$ 0.09          & 3.59 $\pm$ 0.06          & 4.78 $\pm$ 0.07          & 3.45 $\pm$ 0.06          \\
    T-PatchGNN              & 8.01 $\pm$ 0.08          & 4.87 $\pm$ 0.03          & 6.48 $\pm$ 0.09          & 4.19 $\pm$ 0.04          & 4.14 $\pm$ 0.06          & 3.31 $\pm$ 0.04          & 3.69 $\pm$ 0.04          & 3.25 $\pm$ 0.03          \\
    ISTS-PLM                & 8.02 $\pm$ 0.04          & 4.85 $\pm$ 0.08          & 6.43 $\pm$ 0.09          & 4.07 $\pm$ 0.07          & 4.16 $\pm$ 0.04          & 3.19 $\pm$ 0.05          & \underline{3.66 $\pm$ 0.03}    & 3.05 $\pm$ 0.08          \\
    Hi-Patch                & \underline{8.01 $\pm$ 0.06}    & \underline{4.83 $\pm$ 0.04}    & \underline{6.37 $\pm$ 0.05}    & \underline{4.03 $\pm$ 0.09}    & \underline{4.08 $\pm$ 0.02}    & \underline{3.10 $\pm$ 0.07}    & 3.67 $\pm$ 0.02          & \underline{2.97 $\pm$ 0.04}    \\ \midrule
    Ours                    & \textbf{7.63 $\pm$ 0.05} & \textbf{4.70 $\pm$ 0.04} & \textbf{5.83 $\pm$ 0.02} & \textbf{3.82 $\pm$ 0.03} & \textbf{3.67 $\pm$ 0.05} & \textbf{2.89 $\pm$ 0.05} & \textbf{3.34 $\pm$ 0.02} & \textbf{2.86 $\pm$ 0.03} \\ \midrule
    Impr.                   & \multicolumn{2}{c|}{3.68\%}                       & \multicolumn{2}{c|}{6.92\%}                       & \multicolumn{2}{c|}{8.30\%}                       & \multicolumn{2}{c}{6.23\%}                        \\ \bottomrule
\end{tabular}

%% file: table/generality_precision.tex
\begin{tabular}{@{}cc|cccc|cccc|c@{}}
    \toprule
    \multicolumn{2}{c|}{\multirow{2}{*}{Dataset}}            & \multicolumn{2}{c}{NLinear}                       & \multicolumn{2}{c|}{rp ALinear}                   & \multicolumn{2}{c}{iTransformer}                  & \multicolumn{2}{c|}{rp ALinear}                   & \multirow{2}{*}{Impr.}  \\ \cmidrule(lr){3-10}
    \multicolumn{2}{c|}{}                                   & MSE$\times10^{-1}$                & MAE$\times10^{-1}$                & MSE$\times10^{-1}$                & MAE$\times10^{-1}$                & MSE$\times10^{-1}$                & MAE$\times10^{-1}$                & MSE$\times10^{-1}$                & MAE$\times10^{-1}$                &                         \\ \midrule
    \multicolumn{1}{c|}{\multirow{4}{*}{\rotatebox{90}{Electricity}}} & 96  & \textbf{1.97 $\pm$ 0.04} & \textbf{2.75 $\pm$ 0.04} & 1.98 $\pm$ 0.01          & 2.78 $\pm$ 0.02          & 1.51 $\pm$ 0.06          & 2.43 $\pm$ 0.04          & \textbf{1.51 $\pm$ 0.03} & \textbf{2.40 $\pm$ 0.04} & \multirow{4}{*}{0.10\%} \\
    \multicolumn{1}{c|}{}                             & 192 & 2.01 $\pm$ 0.04          & \textbf{2.81 $\pm$ 0.02} & \textbf{2.00 $\pm$ 0.02} & 2.83 $\pm$ 0.02          & 1.61 $\pm$ 0.03          & \textbf{2.56 $\pm$ 0.02} & \textbf{1.61 $\pm$ 0.03} & 2.57 $\pm$ 0.02          &                         \\
    \multicolumn{1}{c|}{}                             & 336 & \textbf{2.11 $\pm$ 0.03} & 3.03 $\pm$ 0.04          & 2.13 $\pm$ 0.02          & \textbf{2.97 $\pm$ 0.02} & \textbf{1.77 $\pm$ 0.04} & 2.71 $\pm$ 0.01          & 1.80 $\pm$ 0.02          & \textbf{2.67 $\pm$ 0.02} &                         \\
    \multicolumn{1}{c|}{}                             & 720 & 2.53 $\pm$ 0.04          & \textbf{3.27 $\pm$ 0.02} & \textbf{2.52 $\pm$ 0.04} & 3.28 $\pm$ 0.03          & 2.18 $\pm$ 0.02          & \textbf{3.09 $\pm$ 0.02} & \textbf{2.14 $\pm$ 0.01} & 3.12 $\pm$ 0.03          &                         \\ \midrule
    \multicolumn{1}{c|}{\multirow{4}{*}{\rotatebox{90}{Traffic}}}     & 96  & 4.61 $\pm$ 0.02          & 3.50 $\pm$ 0.02          & \textbf{4.49 $\pm$ 0.02} & \textbf{3.49 $\pm$ 0.02} & 3.01 $\pm$ 0.03          & 2.57 $\pm$ 0.02          & \textbf{3.01 $\pm$ 0.02} & \textbf{2.54 $\pm$ 0.02} & \multirow{4}{*}{0.38\%} \\
    \multicolumn{1}{c|}{}                             & 192 & 4.61 $\pm$ 0.05          & \textbf{3.51 $\pm$ 0.04} & \textbf{4.59 $\pm$ 0.03} & 3.52 $\pm$ 0.01          & 3.25 $\pm$ 0.03          & 2.75 $\pm$ 0.01          & \textbf{3.21 $\pm$ 0.02} & \textbf{2.72 $\pm$ 0.03} &                         \\
    \multicolumn{1}{c|}{}                             & 336 & 4.89 $\pm$ 0.03          & 3.69 $\pm$ 0.06          & \textbf{4.88 $\pm$ 0.03} & \textbf{3.67 $\pm$ 0.03} & \textbf{3.31 $\pm$ 0.02} & 2.74 $\pm$ 0.02          & 3.33 $\pm$ 0.03          & \textbf{2.73 $\pm$ 0.02} &                         \\
    \multicolumn{1}{c|}{}                             & 720 & 4.93 $\pm$ 0.02          & 3.75 $\pm$ 0.03          & \textbf{4.92 $\pm$ 0.04} & \textbf{3.73 $\pm$ 0.03} & \textbf{3.49 $\pm$ 0.03} & \textbf{2.91 $\pm$ 0.03} & 3.51 $\pm$ 0.02          & 2.93 $\pm$ 0.02          &                         \\ \midrule
    \multicolumn{2}{c|}{Impr.}                              & \multicolumn{4}{c|}{0.25\%}                                                                           & \multicolumn{4}{c|}{0.23\%}                                                                           & 0.24\%                  \\ \bottomrule
\end{tabular}

%% file: author/wcs.tex
obtained his B.D. degree from Beijing University of Posts and Telecommunications in 2022. He is currently a doctoral candidate of State Key Laboratory of Networking and Switching Technology at Beijing University of Posts and Telecommunications. His main research interests include time series analysis, anomaly detection, and multimodal learning.

%% file: author/qq.tex
received the Ph.D. degree from the Beijing University of Posts and Telecommunications, Beijing, China, in 2010. She is currently a Professor with the State Key Laboratory of Networking and Switching Technology, Beijing University of Posts and Telecommunications. Her research interests include edge computing, cloud computing, the Internet of Things, ubiquitous services, deep learning, and deep reinforcement learning.

%% file: author/hb.tex
received the Ph.D. degree from Beijing University of Posts and Telecommunications, China, in 2023. Currently, he is a Postdoctoral Researcher with the State Key Laboratory of Networking and Switching Technology, Beijing University of Posts and Telecommunications. His research interests include 5G/6G networks, multipath networks, collective communication, transmission control, and reinforcement learning.

%% file: author/shf.tex
received the Ph.D. degree from Beijing University of Posts and Telecommunications, Beijing, China, in 2017. He is currently an associate professor with the State Key Laboratory of Networking and Switching Technology, Beijing University of Posts and Telecommunications. His research interests include artificial intelligence, natural language processing, big data analysis, object detection, deep learning, and pattern recognition.

%% file: author/zzr.tex
received the B.S. and Ph.D. degrees from Beijing University of Posts and Telecommunications in 2015 and 2020, respectively. He is currently an associate researcher with the State Key Laboratory of Networking and Switching Technology, Beijing University of Posts and Telecommunications. His research interests involve network routing and management for next-generation network infrastructures using machine learning and artificial intelligence techniques.

%% file: author/wjy.tex
received his Ph.D. degree from the Beijing University of Posts and Telecommunications, Beijing, China, in 2008. He is currently a Professor with the State Key Laboratory of Networking and Switching Technology, Beijing University of Posts and Telecommunications. His research interests include broad aspects of Intelligent Networks, Edge/Cloud Computing, Machine Learning, AIOps and Self-Driving Network, IoV/IoT, Knowledge-Defined Network and Intent-Driven Networking.